\documentclass{article}

\usepackage[preprint,nonatbib]{neurips_2024}

\usepackage[utf8]{inputenc}
\usepackage[T1]{fontenc}
\usepackage{amsmath}
\usepackage{amssymb}
\usepackage{cleveref}
\usepackage{caption}
\usepackage{booktabs}
\usepackage{graphicx}
\usepackage{multirow}
\usepackage{makecell}
\usepackage{array}
\usepackage{boltzbit}

\usepackage{tikz}
\usetikzlibrary{positioning, arrows.meta, fit, backgrounds, calc, shapes.callouts, shapes.geometric}

\usepackage[style=authoryear, maxcitenames=2, maxbibnames=99, giveninits=true, uniquename=false, uniquelist=false, backend=biber]{biblatex}
\crefname{section}{Section}{Sections}
\Crefname{section}{Section}{Sections}
\crefname{table}{Table}{Tables}
\crefname{figure}{Figure}{Figures}
\crefname{equation}{Eq.}{Eqs.}

\title{Infinite-Parameter LLMs: Generating and Adapting Weights from Live Data}

\author{%
Jinli Hu\\
Boltzbit Limited\\
\And
Ross M. Clarke \\
Boltzbit Limited\\
\And
Yichuan Zhang\\
Boltzbit Limited\\
\And
José Miguel Hernández-Lobato\\
University of Cambridge\\
Boltzbit Limited
}

\begin{document}

\maketitle

\begin{abstract}
Scaling laws hold that language models grow more capable with more parameters and more training data. Mixture-of-Experts (MoE) architectures are a remarkable demonstration of these laws, activating only a fraction of an enormous parameter bank for each token. But this success is built on static pretraining data --- the facts and corrections supplied by users during live interactions are a significant untapped source of potential improvement for a deployed model, but cannot be exploited by conventional architectures whose weights are frozen after training. Instead, this newfound knowledge must be placed in the context (by instruction or retrieval) and re-read on every request, only to be discarded afterwards.

We seek instead to learn from live interactions by dynamically updating model weights. Inspired by MoEs, we propose the \textbf{Infinite-Parameter LLM}. A compact hypernetwork turns the online data into low-rank modulations of a shared base network, so feed-forward weights are generated from live data, not read from static memory. Whereas existing weight generators are held fixed after reading the context once, we form a Bayesian belief over the generator's latent state and update it online, such that the effective weights are re-derived as our belief evolves during the session. Although the model's memory footprint is constant, the feasible space of generated weights is thus effectively infinite. Representing live data in the weights rather than the prompt amortises compute, frees the context window, persists updates across turns, and can generalise better than in-context use. Our evaluation protocol applies this methodology to in-context learning and retrieval.
\end{abstract}

\section{Introduction}
\label{sec:intro}

For half a decade, model capability has risen with the amount of training data, alongside parameters and compute \parencite{kaplan_scaling_2020, hoffmann_training_2022}; data is a first-class component of model performance. Two key observations colour this contribution. Firstly, the supply of static textual data is finite: frontier models are on track to exhaust all public human-generated text between roughly 2026 and 2032 \parencite{villalobos_will_2024}, so improving models by simply gathering more training data is becoming less viable. Secondly (and key to this work), data repositories have not stopped growing so much as changed form. Deployed models, and increasingly the agents built on them, generate an enormous and fast-growing stream of \emph{interaction} data, including users' questions, documents, corrections and agentic outcomes. These real-world, inference-time logs are ideally placed to help a model become more useful to a \emph{particular} user on a \emph{particular} task; it is the next source of gold-standard training data.

Fundamentally, today's models cannot authentically \emph{learn} from this data at test-time: once deployed, they are frozen. Existing approaches to simulating model plasticity keep new data \emph{outside} the weights. Some put it in the context (via prompting \parencite{brown_language_2020} or retrieval augmentation \parencite{lewis_retrievalaugmented_2020}) --- alongside the few-shot examples and system prompts which already specify knowledge a model draws on and behaviour it should follow --- to be re-read for every request and then discarded. Others build \emph{around} the frozen model with agent harnesses \parencite{yaoReActSynergizingReasoning2022}, tool orchestration \parencite{schickToolformerLanguageModels2023}, and external memories \parencite{packerMemGPTLLMsOperating2023}, constructing scaffolding which manages data without ever changing the network. But to genuinely and scalably learn from live interaction, a model needs weights which can efficiently absorb new information at inference-time without forgetting existing knowledge.

Our approach is inspired by Mixture-of-Experts (MoE) models, which store a large bank of expert sub-networks and route each token through only a few of them; examples include DeepSeek-V3 \parencite{deepseekai_deepseekv3_2024}, Mixtral \parencite{jiang_mixtral_2024}, Qwen3 \parencite{qwen_qwen3_2025}, Kimi K2 \parencite{kimiteam2025kimik2openagentic}, and Llama-4 \parencite{Llama4Herd}. Equivalently, we can view MoEs as monolothic networks whose weights vary with the input. For a token $x$, an MoE layer applies the effective weight $W_{\text{eff}}(x) = \sum_i g_i(x)\, W_i$, a combination of stored experts selected by an input-dependent gate, following the conditional-computation construction of \textcite{bengioy_estimating_2013,shazeer_outrageously_2017}. In this sense, an MoE is a dense feed-forward network endowed with \emph{dynamic}, per-token weights $W_\mathrm{eff}$, computed as a function of the input. Two issues remain, however. Firstly, all the base weights $W_i$ must be loaded into memory, because a general linear combination may include arbitrary combinations. Secondly, the $W_i$ are still fixed once training ends: an MoE changes its effective weight for each token, but it cannot change the basis underpinning this variation. We avoid both limitations by generating the weights from a compact network which can keep changing during deployment.

Since the information a network holds is bounded at around two bits per parameter \parencite{allenzhu_physics_2024}, a weight generator cannot increase the quantity of knowledge available for any given forward pass, but it can \emph{curate} the available knowledge to focus only on relevant topics. The facts, instructions or demonstrations conventionally supplied in the context are instead `compiled' into the weights, in the scalable ``hypernetwork as encoder'' style \parencite{charakorn_texttolora_2025}. This weight-encoded knowledge follows a a power law in the generator size, and generalises better out-of-distribution than equivalent context encoding \parencite{dhankhar_scaling_2026}. And because task adaptation occupies subspaces of strikingly low intrinsic dimension \parencite{aghajanyan_intrinsic_2021}, only a compact generator network is needed. This approach amortises compute at training time, frees space in the context window, and persists new information to future turns. It also makes absorbing new data dramatically cheaper than conventional fine-tuning: rather than performing forward \emph{and} backward passes over many gradient steps, compiling the same data into the weights costs a \emph{single} forward pass, resulting in \emph{two to three orders of magnitude} less compute, an \emph{order-of-magnitude} smaller memory footprint, and a stored cost of \emph{tens of megabytes} rather than tens of gigabytes. These advantages only widen as models grow (\cref{sec:efficiency}).

Previous work reads the whole context in a single pass, emits one adapter, then holds it fixed while the model answers \parencite{charakorn_texttolora_2025}. To allow variations for \emph{each} token and and across a \emph{whole} session, we must adapt the generator behaviour. We thus parameterise the generator by a latent \emph{code} and carry a \emph{belief} over it, updated once per turn or per token by an amortised recursive Bayesian filter. The code is thus generated from the live conversation and evolves with time.

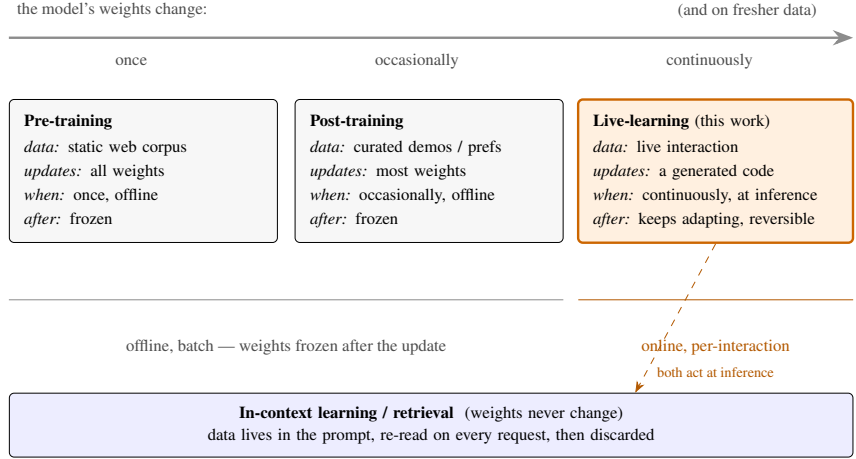
\begin{figure}[t]
  \centering
  \scalebox{0.9}{
\begin{tikzpicture}[
  >=Stealth, font=\footnotesize,
  reg/.style={draw, rounded corners=2pt, align=left, inner sep=2.2mm,
              text width=35mm, fill=black!3, font=\scriptsize},
  ours/.style={draw=orange!80!black, line width=1pt, rounded corners=2pt, align=left,
               inner sep=2.2mm, text width=36mm, fill=orange!12, font=\scriptsize},
  band/.style={draw, rounded corners=2pt, align=center, inner sep=2mm,
               fill=blue!7, font=\scriptsize, text width=120mm},
  ax/.style={font=\scriptsize, text=black!70},
]

\draw[-{Stealth[length=3mm]}, black!45, line width=1pt] (0,4.5) -- (12.41,4.5);
\node[ax, anchor=west] at (0,4.9) {the model's weights change:};
\node[ax, anchor=east] at (12.0,4.9) {(and on fresher data)};
\node[ax] at (1.8,4.15) {once};
\node[ax] at (6.0,4.15) {occasionally};
\node[ax] at (10.3,4.15) {continuously};

\node[reg, anchor=north west] (pre) at (0,3.6) {
  {\bfseries Pre-training}\\[3pt]
  \emph{data:} static web corpus\\[2pt]
  \emph{updates:} all weights\\[2pt]
  \emph{when:} once, offline\\[2pt]
  \emph{after:} frozen};
\node[reg, anchor=north west] (post) at (4.2,3.6) {
  {\bfseries Post-training}\\[3pt]
  \emph{data:} curated demos / prefs\\[2pt]
  \emph{updates:} most weights\\[2pt]
  \emph{when:} occasionally, offline\\[2pt]
  \emph{after:} frozen};
\node[ours, anchor=north west] (live) at (8.35,3.61) {
  {\bfseries Live-learning} (this work)\\[3pt]
  \emph{data:} live interaction\\[2pt]
  \emph{updates:} a generated code\\[2pt]
  \emph{when:} continuously, at inference\\[2pt]
  \emph{after:} keeps adapting, reversible};

\coordinate (bot) at ($(pre.south west |- pre.south)+(0,-3mm)$);
\draw[black!45] (0,0.65) -- (8.15,0.65);
\node[ax, anchor=north] at ($(pre.south west |- bot)!0.5!(post.south east |- bot)+(0,-10mm)$)
     {offline, batch --- weights frozen after the update};
\draw[orange!70!black] (8.37,0.65) -- (12.41,0.65);
\node[ax, anchor=north, text=orange!70!black] at ($(live.south |- bot)+(0,-10mm)$)
     {online, per-interaction};

\node[band, anchor=north west] (icl) at ($(pre.south west |- bot)+(0,-19mm)$) {
  {\bfseries In-context learning / retrieval}\ \ (weights never change)\\[1pt]
  data lives in the prompt, re-read on every request, then discarded};
\draw[->, orange!70!black, dashed] (live.south) -- ($(icl.north)+(3.0,0)$);
\node[ax, anchor=south west, text=orange!70!black, font=\tiny] at ($(icl.north)+(3.2,0.1)$)
     {both act at inference};

\end{tikzpicture}}
  \caption{Three regimes for turning data into model capability, ordered by frequency of weight updates and recency of source data. \textbf{Pre-training} and \textbf{post-training} (SFT, RLHF) both update weights offline and leave them frozen thereafter; they differ mainly in the data they use and how often they run. \textbf{Live-learning} (our work) continuously updates a generated low-rank code at inference, using live interaction data --- facts, corrections and outcomes --- that the other approaches cannot exploit. In-context learning and retrieval (bottom) also act at inference, but fix the weights unchanged and carry data in the prompt, where it is re-read every request and then discarded. The regimes are complementary: live-learning exists alongside pre- and post-training (\cref{sec:experiments}), incorporating data otherwise left on the table.}
  \label{fig:regimes}
\end{figure}

The result is a model with fixed memory footprint and varying effective weights. We call it an `infinite-parameter' LLM, where the term precisely and narrowly references the unbounded set of realisable effective weights and behaviours. By comparison, a conventional model's reachable behaviours are restricted by the fixed weights plus a bounded context. \Cref{fig:regimes} reviews methods of converting data into capability; \Cref{fig:axes,sec:positioning} then review the prior designs inspiring our architecture.

By actively incorporating data generated during model use, the model becomes more useful on the task at hand as an interaction proceeds. Our adaptations are bounded, low-dimensional, and reversible (\cref{sec:moving}), and do not repeal the capacity law or substitute for pretraining. Our evaluation (\cref{sec:experiments}) studies whether carrying data in the weights beats carrying it in the prompt, at matched budget.

We situate the proposal along two orthogonal axes: where the weight comes from, and whether it can change after training. Standard MoE and the bank-free variants ($\mu$MoE, $\infty$-MoE) \emph{select} from a set which, bounded or not, is fixed at deployment. Existing weight generators \emph{generate} weights from context, but read the context once and then freeze the adapter for the turn. We present an approach which is \emph{both generated from live data and adapted online}, making our contribution the coupling of the following elements:

\begin{enumerate}
  \item \textbf{A generative expert architecture and its design space} (\cref{sec:framework,sec:moving,sec:choice}). We make a shared base MLP's weights dynamic through a generated low-rank delta driven by a latent code, with no stored parameter bank, and set out a design space that compares $\infty$-MoE, $\mu$MoE, DFC, HyperMoE, and MoEGen with this structure.
  \item \textbf{The infinite-parameter view} (\cref{sec:infinite}). The precise sense in which the expert space is unbounded and distinguished from unbounded \emph{knowledge}, and a guiding analogy with Bayesian-nonparametric mixtures of experts.
  \item \textbf{Online adaptation as a belief over the latent code} (\cref{sec:moving}). Rather than reading the context once and fixing the adapter, we carry a belief over the latent code and update it as the interaction proceeds, at three cadences (contextual, per-turn, per-token). We incorporate features of uncertainty-gated retention, in-context learning, one-shot hypernetworks, continual-learning posteriors, and fast weights.
  \item \textbf{An evaluation protocol} (\cref{sec:experiments}) comparing encoding information in the weights and carrying them in the prompt. Baselines include in-context learning and retrieval, one-shot weight generators, point-estimate test-time training and vanilla MoE approaches.
\end{enumerate}

\section{Related Work}
\label{sec:related}

Our proposal touches several mature literatures; we organise them below and state our position against the closest work in \cref{sec:positioning}. We claim none of the individual ingredients in isolation.

\subsection{Mixture-of-Experts: from stored banks to bank-free selection}
\label{sec:moe}

The mixture-of-experts idea originates with adaptive mixtures of local experts and their hierarchical, EM-trained form \parencite{jacobs_adaptive_1991, jordan_hierarchical_1994}, and the underlying principle of conditional computation, activating input-dependent parts of a network for capacity without proportional cost \parencite{bengioy_estimating_2013, bengioe_conditional_2015}. Under the scaling-law paradigm, where capacity reliably buys capability \parencite{kaplan_scaling_2020, hoffmann_training_2022}, this made sparsity the default route to cheap capacity: an MoE layer applies a per-token effective weight $W_{\text{eff}}(x)=\sum_i g_i(x)W_i$, a dense FFN whose weights are chosen conditionally on the input. Sparsely-gated MoE realised this at scale \parencite{shazeer_outrageously_2017, fedus_switch_2022}, with subsequent work pursuing finer-grained experts and an always-on \emph{shared expert} (DeepSeekMoE; \cite{dai_deepseekmoe_2024}), a design that directly parallels our always-applied shared base FFN, and very large expert counts via retrieval (PEER; \cite{hex_mixture_2024}) built on product-key memory \parencite{lample_large_2019} and related memory layers \parencite{berges_memory_2024}, with the returns to sparsity themselves the subject of MoE scaling laws \parencite{clark_unified_2022, krajewski_scaling_2024, abnar_parameters_2025}. All of these \emph{store} their experts. Dense-to-MoE ``upcycling'' makes this explicit, replicating a dense FFN into a stored bank \parencite{komatsuzaki_sparse_2023}, the replicate-and-store move we invert. Softer relaxations reduce discreteness but not storage: soft merging of stored experts (SMEAR; \cite{muqeeth_soft_2023}), scaled to autoregressive pre-training (Lory; \cite{zhong_lory_2024}). A separate line reaches an \emph{unbounded but frozen} expert set without a stored bank: $\mu$MoE \parencite[\cref{sec:compress}]{oldfield_multilinear_2024} factorises a fixed weight tensor, and $\infty$-MoE \parencite{takashiro_moe_2026} draws a per-token continuous latent code from a Gaussian router and uses it as a top-$N\%$ \emph{activation mask} over one shared FFN. $\infty$-MoE shares with us a shared base steered by a low-dimensional per-token latent code, but the resemblance is superficial. Its operator is a \emph{multiplicative mask} that only reweights the existing neurons of a vanilla (non-gated) FFN, essentially giving an old-style FFN a GLU-like gate, so on a modern SwiGLU base, which already gates multiplicatively, the mechanism is largely redundant with the architecture. We instead generate an \emph{additive low-rank delta} that steers neurons along new pre-activation directions, drive it from live data rather than a fixed router, and, the difference with no analogue in a frozen router, adapt the latent code online. We treat $\infty$-MoE and $\mu$MoE as the frozen-selection contrast, not as the precedent our method extends.

\subsection{Compressing and factorising experts}
\label{sec:compress}

A large literature makes experts cheaper. Low-rank or vector experts over a shared base recover most of full-expert quality at a fraction of the parameters (MoV/MoLORA, \cite{zadouri_pushing_2023}; MixLoRA, \cite{lid_mixlora_2024}; MoLE, \cite{wux_mixture_2024}; X-LoRA, \cite{buehler_xlora_2024}; LoRAMoE, \cite{dou_loramoe_2024}), but retain a stored bank. Multilinear MoE ($\mu$MoE; \cite{oldfield_multilinear_2024}) is the closest ``do not store experts'' precedent: it represents the whole bank as a single CP- or Tensor-Ring-factorised weight tensor that is never materialised, routed by a differentiable entmax gate, so, like us, it stores no individual experts. The difference is that $\mu$MoE \emph{factorises a fixed tensor and routes linearly over it}, confining each token's effective weight to the convex hull of a fixed atom set, whereas we \emph{generate} the factors from a latent code that is itself produced from live data (\cref{sec:framework}); it is also not adaptable. Because $\mu$MoE already achieves an un-materialised bank, we do not rest our contribution on the absence of storage but on generating the code from run-time data and adapting it online. Orthogonally, resident memory is reduced by quantising and decoding experts on the fly (QMoE; \cite{frantar_qmoe_2023}), pruning or skipping experts \parencite{lu_not_2024}, merging them (HC-SMoE, \cite{chen_retrainingfree_2025}; MEO, \cite{hes_merging_2023}), offloading \parencite{eliseev_fast_2023}, or distilling an MoE into a dense model \parencite{xue_one_2022}, as surveyed by \textcite{liuj_survey_2024}. These compress a \emph{stored bank}; we remove the bank and generate experts instead.

\subsection{Hypernetworks and generated experts}
\label{sec:hypernet}

Hypernetworks generate a target network's weights \parencite{ha_hypernetworks_2017}; more generally, the dynamic-weight-tensor view treats any layer whose weights are an input-dependent function, made tractable by CP factorisation of the generated tensor (DFC; \cite{babiloni_factorized_2023}), the general form our \cref{sec:framework} specialises to the FFN. Lightweight conditioning primitives such as FiLM \parencite{perez_film_2018} and (IA)\textsuperscript{3} \parencite{liuh_fewshot_2022} modulate a shared computation from an input-dependent signal, the same family as an additive or multiplicative weight modulation from a code. The most direct precedents generate a PEFT module for a frozen LLM in a single forward pass over a context or task description (HyperTuning, \cite{phang_hypertuning_2023}; Text-to-LoRA, \cite{charakorn_texttolora_2025}; Doc-to-LoRA, \cite{charakorn_doctolora_2026}; SHINE, \cite{liuy_shine_2026}; Drag-and-Drop LLMs, \cite{liang_draganddrop_2025}; Zhyper, \cite{abdalla_zhyper_2025}), and the concurrent injection-scaling work \parencite{dhankhar_scaling_2026} shows this route \emph{scales}: knowledge injected into a generated adapter improves as a power law in the hypernetwork's size and generalises out of distribution better than a stored LoRA or full fine-tuning. This line is the closest to ours and the one we build on: it establishes that a hypernetwork acting as an \emph{encoder of live data} injects knowledge that the model then uses without the data in context, the ``knowledge from data, not from a bigger bank'' leg of our design. These generators differ sharply in how they read the data, and at what cost. At one end, Text-to-LoRA reads only a short \emph{task description} into a single embedding and generates the adapter from a small MLP, adding well under a percent of the base's parameters \parencite{charakorn_texttolora_2025}. At the other, SHINE reads the \emph{full context} by reusing the frozen backbone itself as the encoder, appending learnable memory tokens processed under an auxiliary ``Meta LoRA'' and mapping their all-layer hidden states to the adapter through a dedicated memory-to-parameter transformer; this reads context far more richly but adds on the order of a sixth of the base's parameters (roughly $(L'/L + 2r/H)\,P$, about $17\%$ for their Qwen3-8B setting; \cite{liuy_shine_2026}). This span, from a compact description-encoder to a backbone-reusing context-encoder, sets the sizing question our own encoder faces (\cref{sec:moving}). The gap we close is orthogonal to it: all of these read the context \emph{once} and then \emph{freeze} the adapter, so the generated weight is turn-level and memoryless, unchanged as the model reads on and reset from one turn to the next (SHINE's recurrent variant chunks a long context but still produces a fixed adapter, not an evolving one). We keep the encoder-of-data generator and add what it lacks, a belief over the latent code that keeps moving as the interaction proceeds (\cref{sec:moving}). Within MoE, HMoE \parencite{qu_hmoe_2022} and HyperMoE \parencite{zhao_hypermoe_2024} generate expert modulations from a low-dimensional latent code but retain the stored bank; \textcite{zhao_hypermoe_2024} report that conditioning the generator directly on the token can underperform a standard MoE, the optimisation difficulty our compact latent bottleneck (\cref{sec:moving}) targets. The effort closest to our generator (see \cref{sec:positioning}) is MoEGen \parencite{zengy_moegen_2026}, which generates instance-specific LoRA updates from a shared hypernetwork, though on the attention projections and from a per-prompt, top-$k$ discrete latent code, without online adaptation. A documented failure mode across weight generators is memorisation rather than generalisation \parencite{zengb_generative_2025}, which we treat as a first-class evaluation concern (\cref{sec:experiments}).

\subsection{Inference-time adaptation and fast weights}
\label{sec:inference}

Adapting weights at inference descends from fast-weight programmers \parencite{schmidhuber_learning_1992, ba_using_2016, schlag_linear_2021}. Test-time training updates weights by self-supervised gradient steps, as a sequence primitive \parencite{sun_learning_2024}, a long-term memory \parencite{behrouz_titans_2025}, reinforcement-learned self-edits \parencite{zweiger_selfadapting_2025}, or per-task adapters that surpass in-context learning on novel structure \parencite{akyurek_surprising_2024}; all produce point estimates. Methodologically, these approaches embed an updatable state \emph{within the sequence-mixing layer} and update it by a hand-designed gradient or ``surprise'' rule; we instead leave attention unchanged, adapt only the generated FFN experts, and replace the hand-designed update with an amortized approximation to an explicit Bayesian filter (\cref{sec:moving}), yielding calibrated retention rather than a point estimate. Any per-token belief we carry is a low-dimensional FFN-side filter, adding no recurrent state to the attention/sequence-mixing path. Test-time compute can instead be spent on search or sampling against a verifier \parencite{snell_scaling_2024}, a matched-budget baseline for us. Closest in spirit are online MoE adaptations: continuous rerouting via gradient-updated router-logit deltas (Rewiring Experts; \cite{su_rewiring_2025}) and gradient-free, uncertainty-guided Bayesian adaptation of expert confidence in medical vision-language models (MoBE; \cite{imam_can_2026}). Both adapt the \emph{usage of a fixed expert bank}, not the latent code of a generated manifold.

\subsection{Continual, online, and Bayesian foundations for adaptation}
\label{sec:continual}

Continual and online learning study exactly the problem of updating a model over time without erasing what it knows, the stability--plasticity trade and its failure mode, catastrophic forgetting \parencite{mccloskey_catastrophic_1989, kirkpatrick_overcoming_2017}. Its three families, regularisation (EWC; online EWC in \emph{Progress \& Compress}, \cite{schwarz_progress_2018}), replay (GEM, \cite{lopezpaz_gradient_2017}), and architecture growth (Progressive Networks, \cite{rusu_progressive_2016}), together with distillation-based variants (Learning without Forgetting, \cite{liz_learning_2017}) all target durable adaptation; \textcite{vandeven_three_2019} taxonomise the settings, and recent work carries the problem to LLMs (\cite{wut_continual_2024}; O-LoRA, \cite{wangx_orthogonal_2023}). A complementary line shows that fixed-capacity networks progressively \emph{lose plasticity} under continual updates \parencite{dohare_loss_2024}. We take two things from this literature. The \emph{framing}: our uncertainty-gating is a stability--plasticity controller that spends plasticity where the posterior is uncertain and protects it where confident, so live adaptation increases the \emph{diversity of weight configurations realised over a session} rather than the stored parameter count. The \emph{machinery}: the recursive posterior-as-prior update (below). We differ by relocating this from full-weight, offline, task-sequential training to a low-dimensional, \emph{generated} latent code updated online at inference, forward-only and anchored to base, so adaptation is bounded and reversible rather than a permanent consolidation.

Probabilistic treatments of MoE run from the original mixtures \parencite{jacobs_adaptive_1991, jordan_hierarchical_1994} through Bayesian hierarchical mixtures of experts \parencite{waterhouse_bayesian_1996}, nonparametric infinite MoE via a Dirichlet-process gate \parencite{rasmussen_infinite_2002}, feature-allocation priors with unboundedly many latent features finitely active \parencite{griffiths_indian_2011}, and modern identifiability/convergence theory for softmax gating \parencite{nguyenh_demystifying_2023}. For LLM-scale adaptation, Bayesian posteriors over low-rank adapters are tractable (Laplace-LoRA, \cite{yang_bayesian_2024}; BLoB, \cite{wangy_blob_2024}), and post-hoc structured Laplace has been applied to MoE expert layers (Bayesian-MoE; \cite{dialameh_bayesian_2025}). Our online update is recursive Bayesian filtering (variational continual learning, \cite{nguyenc_variational_2018}; online Laplace, \cite{ritter_online_2018}, building on \cite{kirkpatrick_overcoming_2017}; low-rank extended Kalman filtering, \cite{chang_lowrank_2023}), but applied to the generator's low-dimensional per-layer latent code rather than to full weights or expert selection. Amortizing such a filter, training a recognition network to emit the state update in a forward pass, places us in the deep state-space / amortized-filtering lineage (deep Kalman filters, \cite{krishnan_deep_2015}; structured inference networks, \cite{krishnan_structured_2017}; deep variational Bayes filters, \cite{karl_deep_2017}; Kalman VAEs, \cite{fraccaro_disentangled_2017}), and we distinguish it from Kalman methods used as \emph{training-time optimizers} over weights, whose observation is the loss rather than a predictive likelihood over a latent code (KOALA++; \cite{xia_koala_2025}).

\subsection{Conditioning on run-time data through the prompt}
\label{sec:prompt}

The incumbent way to make a deployed model use run-time data is to place that data in the context. In-context learning conditions a frozen model on instructions or a few demonstrations supplied at inference \parencite{brown_language_2020}, and can be read as implicit Bayesian inference over a latent concept the context selects \parencite{xie_explanation_2021}; retrieval-augmented generation fetches relevant text into the context so the model can draw on knowledge it does not store \parencite{lewis_retrievalaugmented_2020}, with nearest-neighbour language models a non-parametric variant that interpolates an external datastore at the output \parencite{khandelwal_generalization_2020}. Long-context modelling and soft prompt- or prefix-tuning \parencite{lester_power_2021} are further points on the same axis, enlarging or learning the conditioning signal while the model's own weights stay fixed. All of these carry the run-time knowledge and behaviour \emph{in the context}, where it is re-read on every request, competes for a bounded context window, and is discarded when the request ends; agent harnesses and external memories likewise manage this data around a frozen model rather than writing it into one. Our design targets the same goal by the opposite route, compiling that data into the weights, and \cref{sec:experiments} makes in-context learning and retrieval the primary baselines against which the weight-carried alternative is measured.

\subsection{Positioning: how this work differs}
\label{sec:positioning}

\begin{figure}[t]
  \centering
  \scalebox{0.92}{
\begin{tikzpicture}[
  >=Stealth, font=\footnotesize,
  cell/.style={draw, rounded corners=2pt, minimum width=46mm, minimum height=26mm,
               align=center, inner sep=1.5mm, fill=black!2},
  ours/.style={draw=orange!75!black, line width=0.9pt, rounded corners=2pt,
               minimum width=46mm, minimum height=26mm, align=center,
               inner sep=1.5mm, fill=orange!12},
  ctitle/.style={font=\footnotesize\bfseries, align=center},
  meth/.style={font=\scriptsize\itshape, text=black!60, align=center},
  bank/.style={draw, minimum width=2.6mm, minimum height=2.6mm, fill=gray!30, inner sep=0pt},
  gg/.style={draw, rounded corners=1pt, fill=orange!20, font=\scriptsize, inner sep=1pt,
             minimum width=6mm, minimum height=4mm},
  ad/.style={draw, rounded corners=1pt, fill=orange!35, minimum width=4mm, minimum height=4mm, inner sep=0pt},
  axlab/.style={font=\footnotesize\bfseries, text=black!75},
]

\def\cxa{2.7}   
\def\cxb{8.3}   
\def\cya{1.55}  
\def\cyb{-2.35} 

\node[ctitle] at (\cxa, 3.35) {Weight is \emph{frozen}\\once produced};
\node[ctitle] at (\cxb, 3.35) {Weight \emph{evolves online}\\(belief over the code)};

\node[ctitle, rotate=90] at (-1.35, \cya) {Selected from\\a stored bank};
\node[ctitle, rotate=90] at (-1.35, \cyb) {Generated\\from data};

\node[cell] (TL) at (\cxa, \cya) {};
\node[ctitle, anchor=north] at ([yshift=-1mm]TL.north) {select from a fixed bank};
\begin{scope}
  \foreach \i in {0,1,2}{\foreach \j in {0,1}{
     \node[bank] (blk\i\j) at ($(\cxa-0.45,\cya+0.15)+(\i*0.34,-\j*0.34)$) {};}}
  \node[bank, fill=orange!55] at ($(\cxa-0.45,\cya+0.15)+(0.68,0)$) {};
\end{scope}
\node[meth, anchor=south] at ([yshift=1.2mm]TL.south) {discrete MoE, $\mu$MoE, $\infty$-MoE\\{\scriptsize\upshape(all experts resident)}};

\node[cell] (TR) at (\cxb, \cya) {};
\node[ctitle, anchor=north] at ([yshift=-1mm]TR.north) {adapt a fixed bank's \emph{usage}};
\begin{scope}
  \foreach \i in {0,1,2}{\foreach \j in {0,1}{
     \node[bank] (grk\i\j) at ($(\cxb-0.45,\cya+0.15)+(\i*0.34,-\j*0.34)$) {};}}
  \node[bank, fill=orange!55] (grk_a) at ($(\cxb-0.45,\cya+0.15)+(0,0)$) {};        
  \node[bank, fill=orange!22] (grk_b) at ($(\cxb-0.45,\cya+0.15)+(0.68,-0.34)$) {}; 
  \draw[->, gray!70, shorten >=1pt] (grk_a) to[bend left=25] (grk_b);
  \node[meth, anchor=south, font=\tiny] at ($(\cxb+0.15,\cya+0.42)$) {usage shifts};
\end{scope}
\node[meth, anchor=south] at ([yshift=1.2mm]TR.south) {Rewiring, MoBE\\{\scriptsize\upshape(bank stays; routing/usage shifts)}};

\node[cell] (BL) at (\cxa, \cyb) {};
\node[ctitle, anchor=north] at ([yshift=-1mm]BL.north) {generate once, then freeze};
\begin{scope}
  \node[gg] (glo) at ($(\cxa-0.55,\cyb-0.15)$) {$G$};
  \node[ad] (alo) at ($(\cxa+0.55,\cyb-0.15)$) {};
  \draw[->] (glo) -- (alo) node[midway, above, font=\scriptsize] {once};
  \draw[black!55, line width=0.5pt] ($(alo.north)+(-0.06,0.21)$) arc (150:30:0.09);
  \node[draw, fill=black!55, minimum width=2.4mm, minimum height=1.6mm, inner sep=0pt]
       at ($(alo.north)+(0.02,0.12)$) {};
\end{scope}
\node[meth, anchor=south] at ([yshift=1.2mm]BL.south) {Text-to-LoRA, SHINE, MoEGen\\{\scriptsize\upshape(read context once; fixed for the turn)}};

\node[ours] (BR) at (\cxb, \cyb) {};
\node[ctitle, anchor=north, text=orange!75!black] at ([yshift=-1mm]BR.north) {generate \emph{and} adapte (This work)};
\begin{scope}
  \node[gg] (go) at ($(\cxb-0.85,\cyb-0.1)$) {$G$};
  \node[draw, dashed, draw=orange!75!black, ellipse, fill=orange!18,
        minimum width=11mm, minimum height=6mm, font=\scriptsize, inner sep=0pt]
       (bel) at ($(\cxb+0.25,\cyb+0.28)$) {belief $z_t$};
  \draw[->, orange!75!black] (bel) -- (go) node[midway,left,font=\scriptsize]{};
  \node[ad] (ao) at ($(\cxb+0.3,\cyb-0.35)$) {};
  \draw[->] (go) -- (ao);
  \draw[->, orange!80!black, line width=0.6pt] ($(\cxb-0.85,\cyb-0.72)$) -- ($(\cxb+1.1,\cyb-0.72)$)
       node[midway, below, font=\scriptsize, text=orange!80!black] {evolves over the session $t$};
\end{scope}

\begin{scope}[on background layer, xshift=19pt, yshift=-18pt, xscale=1.3, yscale=1.14]
  \draw[-{Stealth[length=3mm]}, black!35, line width=2pt]
       (-2.15, 2.6) -- (-2.15, -1.5)
       node[axlab, rotate=90, midway, above=1mm] {generate, don't store};
  \draw[-{Stealth[length=3mm]}, black!35, line width=2pt]
       (0.6, 4.4) -- (6.2, 4.4)
       node[axlab, midway, above=1mm] {make the weight live in time};
\end{scope}

\end{tikzpicture}}
  \caption{The two architecture axes of the design, and where prior work sits. \textbf{Down} --- where each token's weight \textbf{comes} from: selected from a stored, fully-resident bank (top), or generated on demand from a compact resident generator (bottom); this is the ``generate, don't store'' move, and it buys a fixed footprint. \textbf{Across} --- what happens to the weight \textbf{after} it is produced: frozen once made (left), or carried as a belief over its latent code and updated online (right); this is the axis that makes the weight \textbf{live in time}. Stored-bank MoE ($\mu$MoE, $\infty$-MoE) and the one-shot weight generators (Text-to-LoRA, SHINE, MoEGen) each sit in a single cell; only the bottom-right --- generate the weight from live data \textbf{and} keep a moving belief over the code --- is occupied by this work. ``Infinite parameters'' is the reach this opens up: an unbounded set of \textbf{effective weights and behaviours} across both facts and time, from a fixed resident footprint --- not an unbounded store of knowledge, which the capacity laws forbid and we do not claim (\cref{sec:infinite,sec:experiments}). Attention is unchanged throughout; the bounded-vs-unbounded geometry of a single generated layer is developed in \cref{fig:designspace}.}
  \label{fig:axes}
\end{figure}
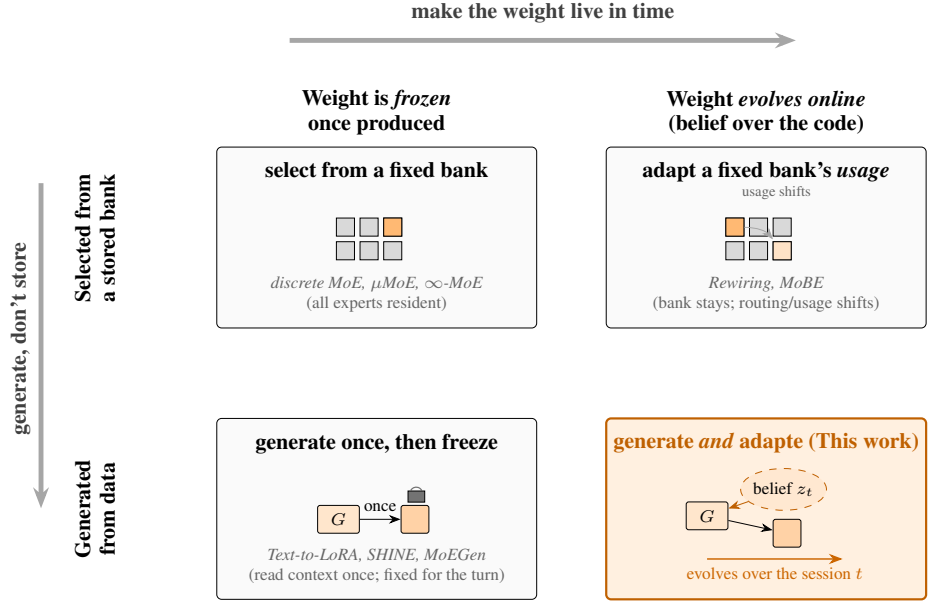

No confirmed prior work combines the full stack we describe, so we position against it on the two axes that survive the reframe: \textbf{where the weight comes from}, selected from a stored bank versus generated from data, and, once generation is granted, \textbf{whether the weight keeps moving after it is produced}, frozen for the turn versus carried as an online-updated belief. Underneath both sits the paradigm contrast that motivates the work, whether run-time knowledge and behaviour are carried in the \emph{weights} or in the \emph{prompt}; the whole generate-and-adapt family lives on the weights side of that line, and in-context learning and retrieval on the prompt side (we treat these as the primary evaluation baselines in \cref{sec:experiments}, not as architectural precedents). \Cref{fig:axes} lays out the two architecture axes and the single cell each prior method occupies; \Cref{tab:lineage} places the closest lineage, the context-driven weight generators, against the axes in detail; and \Cref{tab:adaptation} does the same on the adaptation axis specifically. We give stored-bank MoE only the two-axis summary and not a row-by-row scorecard: as \cref{sec:intro} argued, a stored MoE is the \emph{inspiration} our design departs from and a reference point, not a method we compete with benchmark-for-benchmark, so the detailed comparisons below are with the generator and test-time-adaptation lines that are genuinely close to us.

\begin{table}[t]
  \centering
  \caption{The hypernetwork / generator lineage --- the closest prior work --- against the axes that matter once ``generate rather than store'' is granted. The upper block generates a PEFT module from context in a single pass and then \textbf{freezes} it (turn-level, memoryless); the middle block generates over attention, a stored bank, or the whole weight without a shared low-rank base; the lower block selects from or adapts the usage of a fixed bank. Only this work drives the generator from \textbf{live, accumulating} data and lets the produced weight keep moving, as a calibrated belief over the latent code. ``$\Delta$ over shared base FFN'' marks our specific structure --- an additive low-rank modulation of one always-applied base ($\sim$ = partial: $\infty$-MoE masks a base rather than adding to it); MoBE's posterior is over labels, not weights.}
  \label{tab:lineage}
  \small
  \setlength{\tabcolsep}{4pt}
  \begin{tabular}{@{}p{3.5cm}p{3.3cm}cp{3.2cm}c@{}}
    \toprule
    Work & What drives the generated weight & \makecell{$\Delta$ over\\shared\\base FFN} & Weight after it is produced & \makecell{Post-\\erior} \\
    \midrule
    HyperTuning \parencite{phang_hypertuning_2023} & task / context description & \checkmark & frozen for the turn & $\times$ \\
    Text-to-LoRA \parencite{charakorn_texttolora_2025} & task description & \checkmark & frozen for the turn & $\times$ \\
    Doc-to-LoRA \parencite{charakorn_doctolora_2026} & a document & \checkmark & frozen for the turn & $\times$ \\
    SHINE \parencite{liuy_shine_2026} & in-context prompt & \checkmark & frozen for the turn & $\times$ \\
    Zhyper \parencite{abdalla_zhyper_2025} & conditioning / task & \checkmark & frozen & $\times$ \\
    Injection scaling \parencite{dhankhar_scaling_2026} & fact corpus (train-time) & \checkmark & frozen once baked & $\times$ \\
    MoEGen \parencite{zengy_moegen_2026} & per-prompt discrete code & $\times$ & frozen for the prompt & $\times$ \\
    HyperMoE / HMoE \parencite{zhao_hypermoe_2024, qu_hmoe_2022} & latent code & $\times$ & frozen & $\times$ \\
    DFC \parencite{babiloni_factorized_2023} & raw input & $\times$ & frozen per input & $\times$ \\
    $\mu$MoE / $\infty$-MoE (frozen sel.) & router over fixed atoms & $\sim$ & frozen & $\times$ \\
    Rewiring / MoBE (fixed-bank) & --- (adapts \emph{usage}) & $\times$ & evolves (bank usage) & $\sim$ \\
    \textbf{Inf-params LLMs (This work)} & \textbf{live data + running evidence} & \textbf{\checkmark} & \textbf{evolves online (belief over $z$)} & \textbf{\checkmark} \\
    \bottomrule
  \end{tabular}
\end{table}

Read across these axes, the \emph{generate-instead-of-store} thesis is by now partly anticipated. MoEGen frames the shift from expert \emph{selection} to expert-conditioned \emph{generation}, DFC generates weights from the input in general, the Text-to-LoRA / SHINE line generates adapters from context and shows the route scales, and $\infty$-MoE and $\mu$MoE both reach an un-materialised expert set, so we claim neither that thesis nor the absence of a stored bank as new. The two genuinely unclaimed elements are (i) the \emph{mechanism} as a specific point in the design space of \cref{sec:framework}, a generated low-rank additive delta over a single shared base FFN, driven by a latent code produced from data; and (ii) the \emph{coupling}, in which the generator is driven from \emph{live data} and the latent code it reads is not fixed but carried as a \textbf{belief updated online} by recursive Bayesian inference. The sharpest single distinction is against the generator line closest to us (Text-to-LoRA, SHINE): those read the context \textbf{once and freeze} the adapter, turn-level and memoryless, whereas we carry an evolving belief, so the weight keeps moving within a session. Distillation \parencite{hinton_distilling_2015}, where we use it, is an \emph{enabling} training choice and not a contribution; the reframed design does not rest on compressing a teacher bank. Each rival misses at least one axis: $\infty$-MoE masks rather than generates and is frozen; $\mu$MoE factorises a fixed tensor with linear routing and is frozen; DFC generates a factor but over the input directly, with no shared base or online adaptation; Text-to-LoRA/SHINE generate from context but freeze the adapter; MoEGen generates over attention with a per-prompt top-$k$ code and no online adaptation; Rewiring and MoBE adapt a fixed bank's \emph{usage} rather than a generated latent code.

\begin{table}[t]
  \centering
  \caption{Adaptation positioning: what each method adapts, where, and how. The one-shot weight generators (Text-to-LoRA, SHINE) sit at the top as the closest relatives on the ``generate the weight'' axis --- they produce the adapter from context but freeze it for the turn; the test-time-training methods move a point estimate by gradient descent inside the sequence layer or over the whole model. Ours is the only one to carry a calibrated posterior over a generated latent code, updated online.}
  \label{tab:adaptation}
  \footnotesize
  \setlength{\tabcolsep}{3.5pt}
  \begin{tabular}{@{}p{2.4cm}p{1.9cm}p{1.7cm}p{1.9cm}cp{1.6cm}cc@{}}
    \toprule
    Work & What's adapted & Where it lives & Update rule & \makecell{Pt./\\dist.} & Granularity & \makecell{Unc.} & \makecell{Forg.} \\
    \midrule
    Text-to-LoRA / SHINE & generated LoRA & FFN/attn adapter & read context once & pt. & per-turn (one-shot) & --- & reset each turn \\
    TTT \parencite{sun_learning_2024} & inner-model weights & in sequence layer & gradient & pt. & per-token & --- & implicit \\
    Titans \parencite{behrouz_titans_2025} & memory MLP & branch beside attn. & gradient + momentum & pt. & per-token & --- & gate $\alpha_t$ \\
    SEAL \parencite{zweiger_selfadapting_2025} & full weights & whole model & RL $\to$ SFT & pt. & per-task & --- & --- \\
    Rewiring \parencite{su_rewiring_2025} & router logits & MoE router & gradient & pt. & per-segment & entropy (heur.) & reset \\
    MoBE \parencite{imam_can_2026} & label statistics & frozen experts & gradient-free EMA & post. (labels) & per-sample & \checkmark & --- \\
    \textbf{This work (A--C)} & \textbf{generated latent code $z$} & \textbf{FFN-side, attn. frozen} & \textbf{amortized Bayes filter} & \textbf{dist. over $z$} & \textbf{in-ctx / turn / token} & \textbf{\checkmark (prec.)} & \textbf{\checkmark ($Q$)} \\
    \bottomrule
  \end{tabular}
\end{table}

The adaptation axis tells the complementary story. The closest relatives on the ``generate the weight'' axis, Text-to-LoRA and SHINE, produce the adapter from context but then \textbf{freeze} it for the turn and reset each turn, so the weight does not evolve as the interaction proceeds. The test-time weight-adaptation methods do evolve the weight, but every LLM-side one updates a \emph{point estimate} by gradient descent, inside the sequence-mixing layer (TTT; Titans), over the whole model by reinforcement (SEAL), or over a fixed bank's router logits (Rewiring), while the sole Bayesian one keeps a posterior over \emph{labels}, not parameters, by gradient-free moment-matching (MoBE). None both generates the weight from live data \emph{and} carries a calibrated posterior over the generating latent code, updated by a distilled recursive filter with uncertainty-gating and principled forgetting, on the FFN side with attention untouched.

\paragraph{Concurrent work.} MoEGen \parencite{zengy_moegen_2026} appeared essentially concurrently and independently articulates part of the generate-instead-of-store thesis; we cite it as concurrent, delimit our differences above, and do not claim priority over the shared framing.

\section{Method: Generating and Adapting FFN Experts}
\label{sec:method}

\subsection{Overview}
\label{sec:overview}

\begin{figure}[t]
  \centering
  \scalebox{0.95}{
\begin{tikzpicture}[
  >=Stealth, font=\small,
  base/.style={draw, rounded corners, fill=blue!7, minimum height=40mm, minimum width=20mm, align=center},
  gen/.style={draw, rounded corners, fill=orange!12, minimum height=28mm, minimum width=34mm, align=center},
  box/.style={draw, rounded corners, minimum height=9mm, minimum width=24mm, align=center},
  io/.style={align=center, font=\small},
  ptok/.style={draw=red!75!black, dashed, ->},
  ytob/.style={dashed, ->},
]
  \node[box] (bel) at (2.6, 3.0) {belief $P$\\over $z$};
  \node[box] (z)   at (2.6, 1.1) {latent code $z$};
  \node[box] (dw)  at (2.6,-0.7) {$\Delta$ weights};
  \draw[->] (bel) -- (z) node[midway, font=\scriptsize] {control};
  \draw[->] (z)   -- (dw) node[midway, font=\scriptsize] {proj.};

  \node[base] (base) at (-0.9, 0.4) {Base\\Model};
  \node[io]   (y)    at (-0.9,-2.4) {$y$};
  \draw[->] (base) -- (y);
  \draw[->] (dw) -- (0.1,-0.7) node[midway, above, font=\scriptsize] {apply};   

  \node[gen] (gen) at (7.4, 1.1) {Data-to-Weight LLM\\(dynamic weight generator)};
  \node[io]  (turn) at (7.4,-2.4) {per-turn data};
  \draw[io] (y) -- (turn);
  \draw[->] (turn) -- (gen);
  \draw[->] (gen) -- (z) node[midway, above, font=\scriptsize] {generate};

  \node[io] (x) at (-0.9, 3.0) {$x$};
  \draw[->] (x) -- (base.north);                 
  \draw[ptok] (x) -- (bel) node[midway, above, font=\scriptsize, text=red!75!black, align=left] {amortised per-token\\posterior $q(z\mid x_{1:t})$};

  \draw[ytob] (y.west) |- (-2.5, -2.4) -- (-2.5,4) node[midway, font=\scriptsize, above, align=left, rotate=90] (siglab) {per-token signal}-- (2.6,4) -- (bel.north) ;
\end{tikzpicture}}
  \caption{The architecture, organised around a \textbf{belief over the latent code $z$}. On the main (per-turn) path, live data is read by the \textbf{Data-to-Weight LLM} (the encoder $E_\phi$ of \cref{sec:framework}) into a latent code $z$; the code generates a low-rank weight delta $\Delta W$ that modulates the frozen \textbf{base model}, which produces the output. A \textbf{belief $P$ over $z$} sits above the code and is what makes the weight \textbf{move}: it is updated online, carried from step to step rather than re-encoded from scratch. The per-token signal (dashed) --- the running hidden state, equivalently the realised output, of the autoregressive stream --- feeds the belief and is introduced only to \textbf{amortise the per-token posterior $q(z \mid x_{1:t})$}; it is not on the main data path. The belief's form --- a Gaussian over a continuous code, or a categorical posterior over materialised codes --- is the architectural choice of \cref{sec:choice}.}
  \label{fig:overview}
\end{figure}
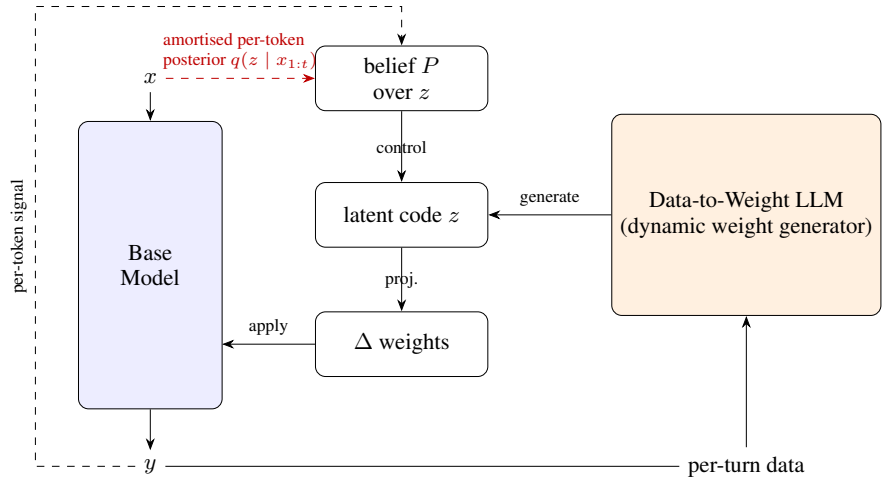

We build on a standard decoder-only transformer and leave attention untouched; only the feed-forward (FFN) sub-layer is changed, and only in a chosen subset of layers. At each such \emph{generative layer}, three components replace the usual FFN (\cref{fig:overview}): a \textbf{shared base FFN}, always applied; a \textbf{compact generator} $G_\phi$ that maps a low-dimensional \emph{latent code} to a structured low-rank modulation of that base; and a \textbf{belief over the latent code}, from which the code driving the generator is read and which is updated from live data and running evidence (\cref{sec:framework,sec:moving}). The token's effective expert is the base FFN plus the generated modulation. Crucially, \textbf{no expert bank is stored}: each token's expert is generated from its latent code and discarded, so the resident parameters are the base, the generator, and the small inference map, all of fixed size, while the set of experts the model can produce is unbounded (\cref{sec:infinite}).

This one mechanism carries all three of our claims. \emph{Infinite parameters:} a fresh expert is generated from a latent code drawn from a continuous, data-materialised space, so the model deploys an unbounded family of \emph{effective weights} rather than reusing a finite stored bank (\cref{sec:infinite}). \emph{Knowledge and behaviour in the weights:} because the code is produced from the data supplied at run time, the effective weights come to carry what a prompt would otherwise carry, such as facts, an instruction, or a few demonstrations, entering through the weights rather than being re-read from the context on every token (\cref{sec:framework}). \emph{Adaptation:} because the expert comes from a latent code, and because we carry a \textbf{belief} over that code rather than reading it once, the model keeps specialising as a session proceeds, with the transformer left unchanged (\cref{sec:moving}). We first fix notation and set out the belief-over-code framework (\cref{sec:framework}); describe how live data writes the belief and how per-token inference moves it (\cref{sec:framework,sec:moving}); make the infinite-parameter claim precise (\cref{sec:infinite}); and only then commit to the concrete architectural choices that instantiate the framework (\cref{sec:choice}).

\subsection{The belief-over-code framework}
\label{sec:framework}

\paragraph{Notation and setup.} We write $d$ for the model width, $h$ for the FFN hidden width, $d_z$ for the latent dimension, and $r$ for the rank of a generated modulation, with $r \ll d$. Layers are indexed by $\ell$ and tokens within a sequence by $t$; the input to a generative layer is the post-attention hidden state $x = h_{\ell,t} \in \mathbb{R}^{d}$, which already integrates context through the layer's attention and the residual stream. The frozen, shared \emph{base FFN} has weights collectively denoted $W_0$ and is initialised from a strong dense model (\cref{sec:setup}). The three moving parts are a \emph{Data-to-Weight encoder} $E_\phi$ that reads run-time data into a latent code, a \emph{generator} $G_\phi$ that maps a code $z$ to a low-rank modulation $\Delta W(z)$ of the base, and a \emph{belief} over the code that is updated online. The code budget $(r, d_z)$, the base architecture, and the concrete values these symbols take are choices we fix in \cref{sec:choice}.

\paragraph{The primitive: a belief over the latent code.} The object at the centre of the design is not a weight and not a code but a \textbf{belief} over the code --- a distribution $P(z)$ that the model carries and updates as it works (we reserve $q$ for the amortized approximation to it, \cref{sec:moving}). Everything else is downstream of it: the code that drives the generator is a summary of the belief (its mean, or its most probable atom), the weight delta is a function of that code, and \emph{adaptation is inference on the belief}. Fixing the belief as the primitive, rather than the weight or a point code, is what lets one mechanism serve the two channels of \cref{sec:intro}: a \textbf{measurement channel}, by which run-time data writes the belief (this is where knowledge and behaviour enter the weights, later in this section), and an \textbf{inference channel}, by which the belief moves between measurements as evidence accumulates (\cref{sec:moving}). The \emph{form} of the belief --- a Gaussian over a continuous code, or a categorical distribution over a pool of materialised codes --- is an architectural choice we defer to \cref{sec:choice}; the framework, and the two channels, are the same either way.

\paragraph{The generated weight.} Given a code $z$, the effective weight of any modulated base projection $W_0$ is the base plus a generated low-rank additive delta,
\begin{equation}
W(z) = W_0 + \Delta W(z), \qquad \Delta W(z) = B(z)\, A(z)^{\top}, \quad A(z) \in \mathbb{R}^{d \times r},\ B(z) \in \mathbb{R}^{h \times r},
\label{eq:generated}
\end{equation}
with the factors $A(z), B(z)$ produced from the code. The delta is never materialised: we compute $y_{\text{delta}} = B(z)\,\big(A(z)^{\top} x\big)$, so the per-token application costs $\mathcal{O}(r\,(d+h))$ per layer, negligible relative to the base FFN's $\mathcal{O}(hd)$. This is the framework; the specific base (SwiGLU), which projections carry a delta, and how $A,B$ are produced are concrete choices made in \cref{sec:choice}. What matters for the framework is only that the weight is a function of a \emph{code}, and the code is drawn from a \emph{belief}.

\paragraph{The measurement channel: writing the belief from live data.} The belief is written from the data supplied for a turn --- the facts, instruction, or examples --- by the Data-to-Weight encoder $E_\phi$ (the encoder-hypernetwork of \cref{sec:hypernet}). This is where the knowledge and behaviour that would otherwise sit in the prompt enters the weights: $E_\phi$ turns supplied data into a code (or, categorically, into a \emph{new materialised code} added to the pool, \cref{sec:choice}), and it is a strong, content-rich measurement rather than a cheap per-token guess. A router over the running hidden state cannot, by itself, inject a fact the base was never given; only the measurement channel can, which is why the encoder and the per-token inference of \cref{sec:moving} are distinct modules.

The encoder is where the design's cost concentrates, and the framework spans a spectrum of realisations trading footprint against reading fidelity; we set out the axis rather than fix a point on it. At the light end, the encoder reads the supplied data with the \emph{base model's own forward pass} (which must process those tokens regardless) and taps a small \emph{readout head} on the resulting hidden states, so the head and generator are the only added parameters, at the scale of a description-conditioned hypernetwork (Text-to-LoRA's smallest variant adds under a percent of the base; \cite{charakorn_texttolora_2025}). In the middle, the backbone is reused as a \emph{dedicated context-encoder} with auxiliary read-time adapters and a memory-to-parameter network, as in SHINE \parencite{liuy_shine_2026}, which reads long context more faithfully at the cost of on the order of a sixth of the base's parameters. At the heavy end, the encoder is a \emph{completely separate hypernetwork}, not tied to the base's weights at all, as in the knowledge-injection hypernetworks of \textcite{dhankhar_scaling_2026}, whose evidence is that injection fidelity scales with this hypernetwork's capacity. These are points on one axis --- how much dedicated machinery reads the data into the code --- and which is warranted is an empirical, footprint-versus-quality question (\cref{sec:selector}) rather than settled by fiat; a relevant consideration along the way is that the online belief (\cref{sec:moving}) can correct an imperfect one-shot read that a frozen generator cannot, which can relieve a lighter encoder of carrying the whole burden in a single pass. Attention and the base weights are frozen throughout.

\subsection{Moving the belief: online inference over the code}
\label{sec:moving}

The element that separates this design from the one-shot weight generators of \cref{sec:hypernet} is that the code is not read once and fixed; we carry the belief and update it as the interaction proceeds. Adaptation, in every variant, is therefore \emph{inference over the latent code}, with the generator and base frozen. The \textbf{transformer's attention / sequence-mixing path is left unchanged throughout}: no variant inserts a recurrent state into the sequence layer, in contrast to test-time-training methods that adapt the sequence path itself \parencite{sun_learning_2024, behrouz_titans_2025}. The only state carried across steps is the low-dimensional belief, and it lives entirely on the FFN side.

\paragraph{Why a belief and not a point.} Existing test-time adaptation carries a \emph{point estimate} of the adapted weights and moves it by gradient descent (TTT, \cite{sun_learning_2024}; Titans, \cite{behrouz_titans_2025}). Carrying instead a posterior over the code earns three things a point cannot. First, \textbf{calibration}: the posterior's spread is an explicit statement of how much to trust the adaptation, usable to gate, abstain, or defer when the model is uncertain. Second, \textbf{uncertainty-gated stability--plasticity}: a precision-weighted update adapts fast where the posterior is unsure and protects what it is confident in, resisting catastrophic forgetting without a bolted-on regulariser (a derived analogue of elastic weight consolidation, \cite{kirkpatrick_overcoming_2017}). Third, \textbf{principled forgetting}: a process-noise term gives a controlled, optionally content-aware way to reopen plasticity when the input distribution shifts. These benefits are carried by the posterior's \emph{spread}, which is exactly the fragile part under amortization, so ``Bayesian'' here is an empirical claim about a calibrated posterior, not a free consequence of emitting a distribution, and validating it --- exact-filter recovery, the amortization gap, and calibration of the posterior precision --- is part of the continuous-Gaussian instantiation we leave to future work (\cref{sec:choice}).

\paragraph{The exact update, and why we amortize it.} The belief is updated by an exact recursive Bayesian filter, the same object at every cadence. At an update, the observation over the tokens since the last update is either the model's own log-likelihood $\mathcal{L}(z) = \sum_i \log p_\theta(x_i \mid x_{<i};\, z)$ (the \emph{self-supervised} regime, always available) or an explicit feedback likelihood $p(y \mid \text{context};\, z)$: Boltzmann in a scalar reward, $p(y \mid z) \propto \exp(r_z / T)$, or Bradley--Terry for a pairwise preference (the \emph{feedback} regime). The recursion is Bayes' rule applied to the running posterior,
\begin{equation}
P_t(z) \;\propto\; P_{t-1}(z)\; \cdot\; p(\text{obs}_t \mid z),
\label{eq:recursion}
\end{equation}
carried from step to step rather than recomputed from scratch. Throughout, we write $P$ for this exact recursive belief and $q$ for the amortized approximation to it that we actually run --- the standard variational reading in which a learned $q$ is fit to a target $P$. Computing the likelihood term exactly requires a \emph{test-time backward pass} to the code, impractical per token at deployment, so the exact belief $P$ is kept only as an offline reference (a distillation teacher, and a comparison baseline) and \emph{amortized}: a trained forward map $F_\phi$ emits the belief update in a single pass, its output $q$ distilled against $P$ \parencite{putzky_recurrent_2017, marino_iterative_2018}; the recognition-network instance of a state-space filter \parencite{krishnan_deep_2015, karl_deep_2017, fraccaro_disentangled_2017}. One property makes a \emph{single} $F_\phi$ serve both cadences below: the exact update over a \emph{window} of tokens is the same function of (prior belief, accumulated observation) whatever the window's length, so $F_\phi$ reads the prior belief and a pooled summary of the window (with a length feature) and is distilled against the exact trajectory at both cadences. This is the \emph{per-token signal} of \cref{fig:overview}, drawn dashed because it exists only to amortise the posterior $q(z \mid x_{1:t})$ --- it is not on the main data path, and switching it off returns the one-shot generator.

\paragraph{The three cadences.} The designs place this one machinery at three points on the belief-granularity axis, indexed by token $t$ or turn $\tau$; they are not three mechanisms but one belief updated more or less often.

\begin{itemize}
  \item \textbf{Design A --- Contextual (implicit belief).} No explicit belief is carried within a sequence; context is integrated by ordinary attention, and a router $R_\ell(h_{\ell,t})$ maps the contextual hidden state to the code. Per-token generation is then an amortized \emph{predictive} inference, the forward pass approximating the Bayesian predictive in-context \parencite{xie_explanation_2021}. This is the cheapest variant and the \emph{degenerate} member of the family --- Bayesian only in the weak sense that in-context learning implicitly approximates a posterior predictive, with none of the calibration, persistence, or controlled forgetting the explicit belief buys. We keep it as the \textbf{baseline} the explicit-belief designs must beat (\cref{sec:accumulation}).
  \item \textbf{Design B --- Session posterior (per-turn update).} An explicit belief is maintained per layer and updated \textbf{once per turn} by $F_\phi$, from the prior belief and a pooled encoding of the turn (and any feedback). Because it fires only per turn, B can equally run the \emph{exact} filter online --- one backward pass per turn is affordable --- making amortization optional here. It gives persistent weight-space adaptation at turn granularity and carries no per-token state.
  \item \textbf{Design C --- Fast belief filter (per-token update).} The belief is carried as a side state and updated \textbf{every token} by the same amortized map, $b_t = F_\phi(b_{t-1}, s_t)$, on a per-token signal $s_t$; the generator reads its summary. Here amortization is essential. This is a genuine per-token weight-space update realised as a \emph{benign, low-dimensional recurrence outside the attention/sequence path}, the finest-grained and fully persistent variant, at the cost of a small carried belief and a cheap forward-only filter step per token.
\end{itemize}

All three instantiate the same idea --- a generated, continuously-indexed expert space adapted by Bayesian inference over its latent code --- and differ only in the granularity and persistence of that inference. The family also locates prior work within one frame: discrete MoE and $\infty$-MoE are the \emph{frozen} limit; in-context learning is the contextual instance (A); and fast-weight/TTT methods are per-token updates placed in the \emph{sequence layer} rather than, as in C, in a low-dimensional FFN-side belief. Because test-time gains tend to accrue with the \emph{number} of updates rather than their size \parencite{sun_learning_2024}, we expect C to dominate B under fine-grained drift, with B the natural read-out when the phenomenon and its labels live at turn granularity; since C run over the whole conversation subsumes B, our accumulation study (\cref{sec:accumulation}) updates at C and reports at the turn level.

\begin{table}[t]
  \centering
  \caption{The three adaptation cadences as one machinery --- a belief over the code updated more or less often. A carries no explicit belief (the baseline); B updates the belief once per turn; C every token. The update rule is the recursive Bayes recursion of \cref{sec:moving} in every case, differing only in the observation window; it is agnostic to the belief's form (the Gaussian or categorical realisations of \cref{sec:choice}).}
  \label{tab:cadences}
  \small
  \begin{tabular}{@{}p{2.6cm}p{3.1cm}p{3.4cm}p{3.7cm}@{}}
    \toprule
     & A --- Contextual & B --- Session (per-turn) & C --- Fast filter (per-token) \\
    \midrule
    Belief update & none (implicit in context) & once per turn $\tau$ & every token $t$ \\
    Carried state & none & belief $P_\tau(z)$ & belief $P_t(z)$ \\
    Update rule & router reads the code & $P_\tau \propto P_{\tau-1}\cdot p(\text{obs}_\tau \mid z)$ & $P_t \propto P_{t-1}\cdot p(x_t \mid z)$ \\
    Wins when & context suffices; short interactions & task shifts across turns; per-turn feedback & long single stream; fine-grained drift \\
    \bottomrule
  \end{tabular}
\end{table}

\subsection{The infinite-parameter view}
\label{sec:infinite}

We call the model an \emph{infinite-parameter} LLM in a precise sense: the set of experts reachable at a generative layer is $\{\,W_0 + G_\phi(z) : z \in \mathcal{Z}\,\}$, where $\mathcal{Z}$ is the space of codes the encoder can \emph{materialise from data}. The stored parameters --- base, encoder, generator, and the small inference map --- are finite and fixed; the reachable effective weights are not, because $\mathcal{Z}$ is not a fixed finite index but a space populated by whatever data the model is given. Over an interaction the model instantiates a growing set of distinct weight configurations rather than reusing a fixed bank.

\begin{figure}[t]
  \centering
  \scalebox{0.95}{
\begin{tikzpicture}[
  >=Stealth, font=\small,
  atom/.style={circle, fill=black, inner sep=1.6pt},
  tok/.style={circle, draw=orange!80!black, very thick, fill=orange!25, inner sep=2pt},
  hull/.style={fill=blue!10, draw=blue!45},
  hulld/.style={draw=gray!55, dashed, thick},
  mfill/.style={fill=orange!12},
  mline/.style={draw=orange!85!black, very thick},
  lbl/.style={font=\footnotesize, align=center},
  capt/.style={font=\scriptsize, align=center},
]
  \begin{scope}[shift={(0,0)}]
    \fill[hull] (0.30,0.40) -- (2.60,0.70) -- (1.30,2.40) -- cycle;
    \node[atom] at (0.30,0.40) {};
    \node[atom] at (2.60,0.70) {};
    \node[atom] at (1.30,2.40) {};
    \node[tok] (s) at (1.35,1.15) {};
    \node[capt, anchor=west] at (1.55,1.15) {$\sum_k c_k(z)\,W^{(k)}$};

    \node[lbl] at (1.3,3.15) {\textbf{selection}\\ (linear routing over static\\ atoms: $\mu$MoE / dictionary / MoE)};
    \node[capt] at (1.3,-0.85) {reachable set $=$ convex hull of a\\ \emph{finite} atom set \; \textbf{(bounded)}};
  \end{scope}

  \draw[dotted] (4.15,-1.1) -- (4.15,3.5);

  \begin{scope}[shift={(5.2,0)}]
    \fill[mfill] (0.30,0.40)
      .. controls (1.45,-0.25) .. (2.60,0.70)
      .. controls (3.35,1.55) .. (1.30,2.40)
      .. controls (-0.05,1.35) .. (0.30,0.40);
    \draw[mline] (0.30,0.40)
      .. controls (1.45,-0.25) .. (2.60,0.70)
      .. controls (3.35,1.55) .. (1.30,2.40)
      .. controls (-0.05,1.35) .. (0.30,0.40);
    \draw[hulld] (0.30,0.40) -- (2.60,0.70) -- (1.30,2.40) -- cycle;
    \node[atom] at (0.30,0.40) {};
    \node[atom] at (2.60,0.70) {};
    \node[atom] at (1.30,2.40) {};
    \node[tok] (g) at (2.45,1.48) {};
    \node[capt, anchor=west] at (2.62,1.62) {$W_0+G_\phi(z)$\\ \emph{outside the hull}};

    \node[lbl] at (1.3,3.15) {\textbf{generation}\\ (nonlinear $G_\phi$ over a\\ continuous latent code $z$: ours)};
    \node[capt] at (1.3,-0.85) {reachable set $=$ a curved manifold in\\ \emph{no} finite-dim.\ affine span \; \textbf{(unbounded)}};
  \end{scope}

  \node[atom] at (-1.55,-1.75) {};
  \node[capt, anchor=west] at (-1.40,-1.75) {teacher / anchor experts};
  \node[tok] at (2.55,-1.75) {};
  \node[capt, anchor=west] at (2.75,-1.75) {expert used for one token};
  \node[capt, anchor=west] at (5.55,-1.75) {\textcolor{gray!70}{-\,-\,-} convex hull (selection's reach)};
\end{tikzpicture}}
  \caption{Selection over a stored bank versus generation over a code space, on the same three anchor experts. \textbf{Left:} routing over a finite set of \textbf{stored} atoms ($\mu$MoE / discrete MoE) reaches only their convex hull (the triangle); every routed expert lies strictly inside it --- bounded. \textbf{Right:} codes \textbf{generated from data} by $E_\phi$ populate a curved manifold that bulges beyond that hull (shown dashed), so a generated expert $W_0 + G_\phi(z)$ can lie strictly \textbf{outside} it --- the reachable set is contained in no finite-dimensional affine span, and is unbounded. This is the geometric content of the infinite-parameter claim: what matters is that the atoms are \textbf{generated} rather than \textbf{stored}, not whether the belief over them is continuous or categorical. This paper's categorical belief is a finite working set over this unbounded generated space --- a moving slice of the right panel, not a return to the left.}
  \label{fig:designspace}
\end{figure}
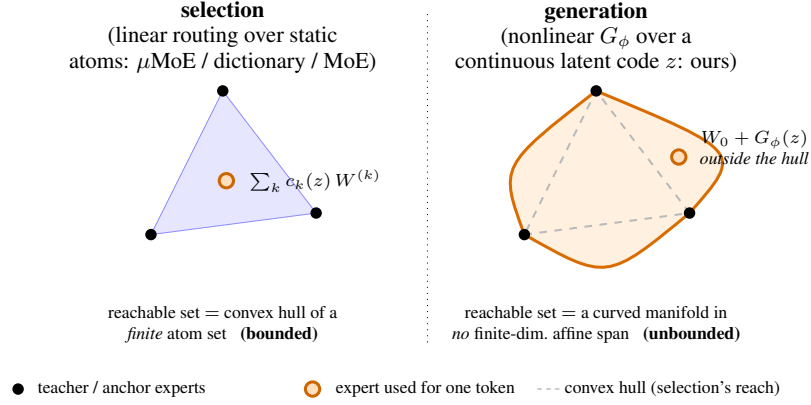

This is where the categorical instantiation of \cref{sec:choice} must be positioned carefully, because it \emph{looks} like the finite selection the paper otherwise argues against. The distinction is the origin of the atoms. A classical MoE selects among a \textbf{fixed, stored} bank of experts; its reachable set is the convex hull of those atoms --- bounded, a \emph{selection} (\cref{fig:designspace}, left). Our categorical belief is a posterior over a pool of atoms that are themselves \textbf{generated from data} by $E_\phi$: any new data materialises a new code, so the pool is unbounded and the atoms are drawn from a continuum, not enumerated in advance. A categorical belief over a data-materialised pool is thus the \emph{finite, tractable working-set representation} of a belief over an unbounded generated space --- the same relationship a Dirichlet-process mixture has to its infinite base measure, where any computation touches only a finite active set while the pool of possible components is unbounded \parencite{rasmussen_infinite_2002}. The unboundedness the name claims therefore does not require a \emph{continuous code at inference}; it requires that codes be \emph{generated rather than stored}, which the measurement channel (\cref{sec:framework}) guarantees. Selection over a \emph{stored} bank is bounded; selection over a \emph{generated} pool is not.

Two clarifications keep the claim honest. First, ``infinite'' is a statement about \emph{reachable weight configurations}, not stored knowledge: knowledge remains bounded by the resident parameters \parencite{allenzhu_physics_2024}, and ``infinite'' here never means a larger knowledge store. Second, adaptation adds no parameters; it \emph{re-allocates plasticity}, since the belief's uncertainty (\cref{sec:moving}) decides which latent directions stay plastic and which are protected, resolving the stability--plasticity trade \parencite{dohare_loss_2024} at inference rather than freezing it. We are careful to claim only what is ours: that a layer's weights can be made a data-dependent function rather than a stored constant is established (hypernetworks, \cite{ha_hypernetworks_2017}; dynamic layers, \cite{babiloni_factorized_2023}), and MoE is itself a dynamic-weight layer with a finite index; our contribution is the specific coupling --- a belief over a \emph{generated} code space, written by live data and moved by online inference --- not dynamic weights in the abstract.

This positioning also separates us from the neighbouring generated- and selected-expert methods along one axis, the origin of the atoms and whether the belief moves: $\mu$MoE \parencite{oldfield_multilinear_2024} and discrete MoE \emph{select} over a stored bank (bounded); DFC \parencite{babiloni_factorized_2023} and MoEGen \parencite{zengy_moegen_2026} \emph{generate} an adapter and freeze it after one read; $\infty$-MoE \parencite{takashiro_moe_2026} masks subsets of one fixed network; and HyperMoE \parencite{zhao_hypermoe_2024} generates a supplementary branch over a stored bank. None carries an online belief over a \emph{data-materialised} pool, which is the coupling this paper adds.

\subsection{Architecture choice in this paper}
\label{sec:choice}

The framework above is deliberately agnostic about the form of the belief and the shape of the generator. We now commit to the choices this paper evaluates: the \emph{form of the belief} (categorical, \cref{sec:categorical}), and the \emph{generator and base} it drives (\cref{sec:generator}). The alternative --- a continuous-Gaussian belief with a nonlinear generator, the framework's most expressive point --- we note as a further direction at the end of this section rather than evaluate here.

\subsubsection{A categorical belief over materialised codes}
\label{sec:categorical}

We instantiate the belief over $z$ as a \textbf{categorical} distribution over a pool of codes $\{m_1, \dots, m_K\}$, each materialised from data by the encoder $E_\phi$. The belief is $P_t(z) = \mathrm{Cat}(\pi_t)$ with $\pi_t \in \Delta^{K-1}$, the code driving the generator is the posterior's most probable atom (top-1) or its mean, and the online update of \cref{sec:moving} becomes recursive categorical Bayes,
\begin{equation}
\pi_{t,k} \;\propto\; \pi_{t-1,k}\; \cdot\; p(\text{obs}_t \mid z = m_k),
\label{eq:categorical}
\end{equation}
so that as the interaction proceeds the belief concentrates on the code that best explains the running evidence, and re-opens when the evidence shifts. This is the exact recursive filter of \cref{sec:moving} specialised to a categorical latent; its amortization is a learned \textbf{selector} that emits the posterior over the pool in a single forward pass. Concretely, the selector scores the running hidden state against each code and normalises: at generative layer $\ell$ with the layer-input activation $u_\ell$ as query and a learned key $\kappa_{\ell}(m_k)$ per code, $\pi \propto \exp\langle u_\ell, \kappa_\ell(m_k)\rangle$. Selection is top-1 per layer, so a single generated expert is applied --- not a top-$k$ mixture --- which keeps the operator a genuine weight rather than an averaged one; the measurement channel of \cref{sec:framework} supplies the codes, and the selector supplies the cheap per-step inference over them. The two cadences of \cref{sec:moving} carry over directly: per-turn (B), the posterior is updated once per turn as questions accumulate over a fixed knowledge pool; per-token (C), it is updated as the sequence streams.

\emph{(The empirical study of this selector --- how well the categorical posterior identifies the code that carries the answer, what signal drives it, and where in the network the routing signal lives --- is the subject of \cref{sec:selector}.)}

\subsubsection{The low-rank generator}
\label{sec:generator}

For the generator and its base we adopt the concrete pipeline of SHINE \parencite{liuy_shine_2026} essentially unchanged, and materialise the categorical pool of \cref{sec:categorical} by running it once per knowledge set. The base is a strong dense SwiGLU model \parencite[fixed in \cref{sec:setup}]{shazeer_glu_2020}, and the generated delta modulates its FFN projections $\{W_{\text{gate}}, W_{\text{up}}, W_{\text{down}}\}$ with a small rank ($r = 8$) and latent dimension ($d_z = 128$), the code-budget controls of the framework.

\paragraph{Reading data into a code.} The base is a frozen decoder-only transformer. To read context, its tokens are passed through the base with a set of $M$ learnable \textbf{memory tokens} appended to the sequence; these are input-independent probes, trained once and shared, that read information out of the evidence by ordinary attention. The memory tokens' hidden states are collected \textbf{from every layer}, giving a memory grid $\mathbf{m} \in \mathbb{R}^{L \times M \times d}$. A \textbf{memory-to-parameter (M2P) network} then mixes this grid, and emits a flat \textbf{latent code} $z \in \mathbb{R}^{P}$, from which a trivial projection applies to give the LoRA parameters the base needs.

\paragraph{Reshaping the code into weight deltas.} For a weight $W_0 \in \mathbb{R}^{\text{out} \times \text{in}}$ the low-rank (LoRA; \cite{hu_lora_2022}) factors $A \in \mathbb{R}^{\text{in} \times r}$, $B \in \mathbb{R}^{\text{out} \times r}$ and an optional bias $C \in \mathbb{R}^{\text{out}}$ are applied as
\begin{equation}
W(z)\,x \;=\; W_0\,x \;+\; \big(\sqrt{s}\,B\big)\big(\sqrt{s}\,A\big)^{\!\top} x \;+\; s\,C,
\label{eq:lora}
\end{equation}
with a fixed scale $s$ folded as $\sqrt{s}$ into each factor and $s$ into the bias. The rank is small ($r=8$), so each adapter is cheap; the memory-token count is set so the flat code $z$ has exactly the size the per-layer LoRA factors require. The delta is applied in factored form, $B(A^{\top}x)$, never materialised, so per-token cost is $\mathcal{O}(r\,(\text{in}+\text{out}))$ per projection.

\paragraph{Why this generator, and what we change.} Two properties make this the right generator for our framework. First, it is a \emph{faithful, high-bandwidth} reader: unlike a compact readout head, the memory-token/M2P path reads long evidence into a code that reconstructs per-layer adapters well enough to answer questions the base was never given (the \cref{sec:d2w} result on which this paper's data-to-weights claim rests). Second, it is \emph{deterministic and cacheable}: one read per knowledge set yields a code, and that code is exactly the materialised atom $m_k$ of the categorical pool (\cref{sec:categorical}). We take the generator, memory tokens, M2P network, and meta-LoRA \textbf{frozen} from a SHINE checkpoint and add only the categorical selector of \cref{sec:categorical} on top; the sole trainable parameters introduced by this paper are the selector's per-code key map and its query alignment, at a scale of well under a percent of the base. The code-to-weight reshape here is \emph{linear} in the code, with the nonlinearity of the read concentrated in the encoder (the memory/M2P stack) rather than the code-to-weight step --- a preliminary finding of ours is that a linear code-to-weight leg ties a nonlinear one at a fraction of the parameters, which is why we adopt it.

\paragraph{Cost.} Codes are computed once per knowledge set and cached, so at run time the only cost beyond a base forward pass is (i) the selector's $K$ inner products per layer to update the categorical belief and (ii) applying the selected code's factored deltas. Both are negligible relative to the base; in particular, nothing re-reads the evidence tokens at generation time. This is the concrete sense in which carrying data in weights, once compiled, is cheaper at run time than re-reading it from the prompt on every token (\cref{sec:d2w}), and it is the property the dilution study of \cref{sec:d2w} exploits when the evidence is too large to keep re-reading in-context.

\paragraph{The richer belief we do not evaluate.} The categorical form chooses \emph{among} whole-code reads rather than moving \emph{within} the code space, so a shift the pool does not already contain can be met only by materialising a new atom. The framework's more expressive point (\cref{sec:framework}) instead carries a \textbf{continuous-Gaussian belief} over a zero-anchored code offset $\psi_\ell \sim \mathcal{N}(0,\Sigma_0)$, $z_{\ell} = c_\ell + \psi_\ell$ (so $\psi_\ell=0$ recovers the un-adapted model), and turns the recursive update of \cref{sec:moving} into a Laplace / extended-Kalman filter whose posterior \emph{precision} gates plasticity --- adapting fast where it is uncertain, protecting what it is confident in, a derived analogue of elastic weight consolidation \parencite{huszar_note_2018, chang_lowrank_2023, kirkpatrick_overcoming_2017}. This is the form in which ``Bayesian'' becomes load-bearing rather than decorative, but it demands a code-to-weight map smooth in $z$, the exact filter as a distillation teacher, and calibration of the amortized precision; we leave it to future work and evaluate the categorical belief here.

\section{Experiments}
\label{sec:experiments}

Our experiments are set up to answer three questions, each resting on the one before and each the subject of one subsection, which together test the design promise that the infinite-parameter LLM can learn from its live interaction by writing that interaction into its weights, and go on adapting as the interaction grows. The first is whether run-time data can enter the weights and be used at all: does compiling a turn's evidence into the generated weight let the model answer from it, with the evidence withheld from the prompt (\cref{sec:d2w})? The second arises once a session has written several pieces of data into a pool of codes --- whether the model can infer \emph{which} of them the current query needs, the single-step form of the belief over the code (\cref{sec:selector}). The third is whether that belief \emph{accumulates} across the interaction, so the model routes better as the conversation lengthens than it would by treating each turn afresh (\cref{sec:accumulation}).

\subsection{Setup}
\label{sec:setup}

\paragraph{Base and generator.} We build on a frozen base (Qwen3-8B; \cite{qwen_qwen3_2025}) and a data-to-weights generator that compiles evidence into low-rank weight deltas (the pipeline of \cref{sec:generator}); the generator is reused from prior work rather than retrained here. On top of this we add the categorical selector of \cref{sec:categorical}, which is trained lightly on a routing objective. Full training details are outside the scope of this paper.

\paragraph{Data and tasks.} We evaluate on five question-answering datasets spanning the axis that matters for weights-versus-prompt --- how long, noisy, and multi-hop the evidence is. \textbf{SQuAD} (single short passage, clean) is the easy end, where the prompt is cheap and strong. \textbf{MS MARCO v2.1} (a question with $\approx 10$ candidate passages, one marked answer-bearing) is the long, noisy, multi-passage end. Between them sit three \textbf{multi-hop} sets whose answers require combining several passages: \textbf{HotpotQA} (distractor setting: 2 gold + 8 distractor paragraphs), \textbf{2WikiMultihopQA}, and \textbf{MuSiQue} (the hardest, built to resist single-hop shortcuts). The multi-passage sets carry per-passage gold relevance labels (\texttt{is\_selected} in MS MARCO, supporting-fact annotations in the multi-hop sets), which give the selector experiment (\cref{sec:selector}) a routing target for free; SQuAD, having a single passage, is used only for the weights-versus-prompt comparison. Unless noted, results are over $n=150$ held-out groups, scored by answer F1 (generation) or top-1/top-3 routing accuracy (selection).

\paragraph{Baselines.} For weights-versus-prompt: \emph{closed-book} (no evidence), \emph{in-context} (evidence in the prompt), and the \emph{one-shot data-to-weights} read. For selection over the code pool: \emph{random} ($1/K$), \emph{BM25} and \emph{dense retrieval} (bge-small, untrained) over the same candidate passages --- the standard, strong way to pick the right passage --- and an \emph{oracle} that scores each code by the likelihood it assigns the true answer, which upper-bounds the routing signal.

\subsection{Data-to-weights beats the prompt where evidence is long and multi-hop}
\label{sec:d2w}

We first \textbf{reproduce} the data-to-weights generator we build on (SHINE; \cite{liuy_shine_2026}) on our own setup, to confirm on a validated base that run-time evidence compiled into the weights can actually be used. We then run a \textbf{dilution study}, new here, that probes \emph{where} the single one-shot read breaks as evidence scales, and that motivates per-token dynamic adaptation.

\paragraph{Reproducing the base: data-to-weights versus the prompt.} Whether compiling a turn's evidence into the code beats carrying it in the prompt depends entirely on the evidence (\cref{tab:weightsvsprompt}). On \textbf{SQuAD} --- one short, clean passage --- the prompt is the ceiling (in-context 85.3 vs data-to-weights 51.8): when the evidence is small and used once, nothing beats simply reading it. On \textbf{MS MARCO} --- ten passages, mostly distractors --- the picture inverts: data-to-weights reaches 48.0 F1 against the in-context 33.6, because the prompt now pays for length and noise while the compiled code does not. The three multi-hop sets sit on the weights-favoured side of the crossover, and are the datasets that most sharply test the claim: the answer spans several passages, so the prompt must hold them all while the code compiles them.

\begin{table}[t]
  \centering
  \caption{Weights versus prompt across the evidence-difficulty axis (measured in F1). The prompt wins when evidence is short and clean (SQuAD); compiling into weights wins when it gets longer and noisier (the others).}
  \label{tab:weightsvsprompt}
  \begin{tabular}{@{}lrrr@{}}
    \toprule
    Dataset (evidence) & Closed-book & In-context & Data-to-weights \\
    \midrule
    SQuAD (1 short passage) & 20.2 & \textbf{85.3} & 51.8 \\
    HotpotQA (2-hop, +distractors) & 22.1 & 58.7 & \textbf{60.4} \\
    2WikiMultihopQA (multi-hop) & 24.5 & 55.5 & \textbf{58.1} \\
    MuSiQue (hard multi-hop) & 15.2 & 40.9 & \textbf{45.3} \\
    MS MARCO v2.1 (10 passages) & 16.8 & 33.6 & \textbf{48.0} \\
    \bottomrule
  \end{tabular}
\end{table}

\begin{figure}[t]
  \centering
  \scalebox{0.95}{
\begin{tikzpicture}[>=Stealth, font=\small]
  \draw[->] (0,0) -- (8.4,0) node[right, font=\footnotesize] {pool size (answer + $k$ distractors)};
  \draw[->] (0,0) -- (0,5.2) node[above, font=\footnotesize] {answer F1};
  \foreach \yy/\lab in {1.167/30, 2.7/40, 4.233/50}{
    \draw[gray!25] (0,\yy) -- (8.4,\yy);
    \node[left, font=\scriptsize, gray] at (0,\yy) {\lab};
  }
  \foreach \x/\lab in {0.6/0, 2.4/8, 4.0/16, 5.6/32, 7.4/64}{
    \node[below, font=\scriptsize] at (\x,0) {\lab};
  }
  \draw[very thick, blue!70!black] (0.6,2.7) -- (2.4,4.479) -- (4.0,3.957) -- (5.6,4.356) -- (7.4,4.019);
  \node[blue!70!black, font=\scriptsize, right] at (7.4,4.10) {oracle @1300};
  \draw[very thick, blue!70!black, dashed] (0.6,2.7) -- (2.4,4.479) -- (4.0,4.019) -- (5.6,3.758) -- (7.4,3.175);
  \node[blue!70!black, font=\scriptsize, right] at (7.4,3.175) {oracle @3000};
  \draw[very thick, orange!85!black, dashed] (0.6,2.7) -- (2.4,4.049) -- (4.0,3.559) -- (5.6,2.961) -- (7.4,1.841);
  \node[orange!85!black, font=\scriptsize, right] at (7.4,1.90) {realistic @3000};
  \draw[very thick, orange!85!black] (0.6,2.7) -- (2.4,4.049) -- (4.0,3.559) -- (5.6,1.933) -- (7.4,0.829);
  \node[orange!85!black, font=\scriptsize, right] at (7.4,0.829) {realistic @1300};
  \foreach \p in {(0.6,2.7),(2.4,4.479),(4.0,3.957),(5.6,4.356),(7.4,4.019)}{\fill[blue!70!black]\p circle(1.1pt);}
  \foreach \p in {(0.6,2.7),(2.4,4.479),(4.0,4.019),(5.6,3.758),(7.4,3.175)}{\fill[blue!70!black]\p circle(1.1pt);}
  \foreach \p in {(0.6,2.7),(2.4,4.049),(4.0,3.559),(5.6,2.961),(7.4,1.841)}{\fill[orange!85!black]\p circle(1.1pt);}
  \foreach \p in {(0.6,2.7),(2.4,4.049),(4.0,3.559),(5.6,1.933),(7.4,0.829)}{\fill[orange!85!black]\p circle(1.1pt);}
\end{tikzpicture}}
  \caption{The dilution boundary (MS MARCO v2.1, top-1 answer F1, measured, $n=150$), at two encoder context budgets (1300 solid, 3000 dashed). \textbf{Oracle} (blue) keeps the answer passage fronted so it survives truncation; \textbf{realistic} (orange) shuffles passage order so the answer is as exposed as any other. Even the oracle declines as the pool grows with nothing truncated ($51.6 \to 46.9$ F1 from 8 to 32 distractors at 3000 tokens, 0\% truncation) --- the code \textbf{saturates}: a fixed-size code loses fidelity as it carries more. The realistic read falls further below the oracle because the answer is either truncated out (dominant at 1300 tokens, where the 32- and 64-passage reads are 100\% truncated) or, once the budget is large enough that nothing truncates (0\% at 3000 for $\le 32$ passages), simply buried among distractors and read less faithfully. Saturation bounds how much one code should hold; truncation and burial are failures of foregrounding the right evidence --- together they motivate one bounded read per code plus a selector over the pool, rather than one ever-larger read.}
  \label{fig:dilution}
\end{figure}
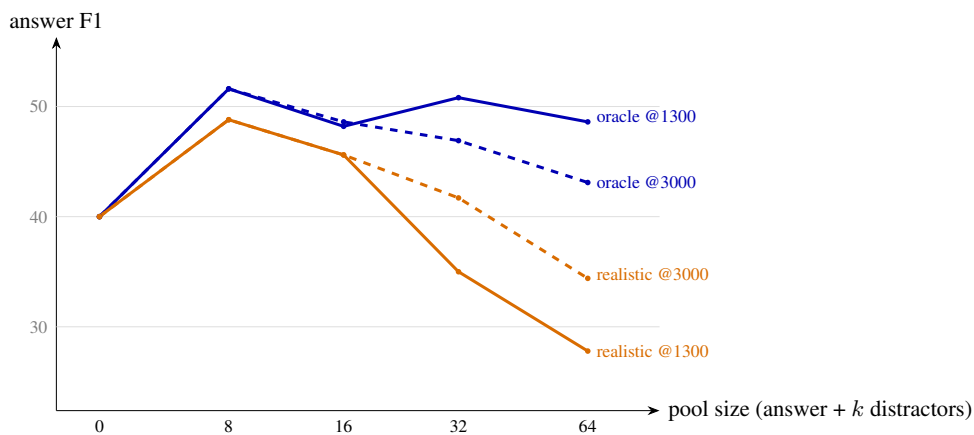

\paragraph{The dilution boundary.} Does a fixed-size code \emph{dilute} as more evidence is packed into it? We hold the answer-bearing passage in the pool, add up to 64 distractor passages, and compare two placements: \textbf{oracle} (the answer passage kept at the front, so it survives) and \textbf{realistic} (passage order shuffled, so at inference --- where the model does not know which passage carries the answer --- it is as exposed as any other). We run this at two encoder context budgets, 1300 and 3000 tokens, to separate the effect from any one window size (\cref{fig:dilution}).

Two effects stand out, and the two budgets separate them. First, the code \textbf{does} saturate: even the oracle placement, with the answer passage fronted and nothing truncated, declines as the pool grows --- at the 3000-token budget it falls $51.6 \to 48.6 \to 46.9$ F1 from 8 to 32 distractors with truncation held at 0\%, so a fixed-size code genuinely loses fidelity as it is asked to carry more, independent of where the answer sits. Second, on top of saturation, the realistic placement falls \emph{further} below the oracle, and \emph{why} it falls further has two causes the budgets tease apart. At the small budget the answer passage is \textbf{truncated} out of the window as the pool overflows (at 1300 tokens, 100\% of reads truncate by 32 passages and realistic F1 collapses to 27.8). Raising the budget to 3000 pushes that cliff back --- but does not close the oracle--realistic gap: at 32 passages \emph{nothing} is truncated (0\% at 3000) and yet the realistic read still trails the oracle by $\approx$5 F1, because a \textbf{buried} answer passage is read less faithfully than a fronted one even when both fully fit. The three effects compound, but they divide into one about capacity and two about foregrounding. Saturation is a real cost of any single read, and bounds how much one code should be asked to hold. Truncation and burial are instead failures of \emph{which} evidence the read spends its budget on, because at inference it does not know which passage carries the answer. The oracle--realistic gap --- $\approx$8--20 F1 depending on budget --- is the value on the table for a mechanism that can \emph{identify} the right evidence rather than commit to one fixed read, and the saturation curve is the reason not to answer that by simply reading more into one code. This motivates carrying a belief over a pool of pre-encoded codes and sharpening it dynamically (\cref{sec:selector,sec:accumulation}): each code reads one bounded passage in-window offline, small enough to stay clear of saturation. The question is then no longer \emph{what fits, or sits first, in one read} but \emph{which code the belief selects} and, across a session, \emph{how that selection improves as evidence accumulates}.

\subsection{A trained belief over the code pool beats retrieval}
\label{sec:selector}

Given one pre-encoded code per candidate passage, we ask whether a belief over the pool can route a question to the code carrying its answer. The routing signal is real but not free (\cref{tab:routing}): on MS MARCO, an oracle that scores each code by the likelihood it assigns the \emph{true} answer routes almost perfectly (78.7 top-1, 96.7 top-3), confirming the codes are strongly separable --- but a zero-shot proxy that scores each code by the model's confidence in its own answer is near-random (22.7), so the belief must be \emph{trained}, not read off for free.

Trained, the activation-routed selector (a query taken from the base's own layer activations, scored against a learned key per code, \cref{sec:categorical}) routes far above random and lexical baselines and \textbf{beats dense retrieval over the same candidates on every dataset}, by 8 F1 on MS MARCO (53.3 vs 45.3) and 10--12 on the multi-hop sets (e.g. 70.1 vs 58.1 on 2Wiki, 53.0 vs 40.9 on MuSiQue). Retrieval is the honest bar here --- it, too, picks the right passage --- so beating it establishes that a belief over the generated codes, read from the base's own activations, carries \emph{more} single-question routing signal than a strong text retriever, while operating over compiled codes rather than re-read passages. Two findings from the MS MARCO runs explain where the signal comes from: it lives in the network's \textbf{later layers} (early-layer activations route near-random, late-layer ones carry almost all of it), and taking the query from a single late-layer summary (token-0 of the code, \cref{sec:generator}) outperforms pooling all memory tokens --- the routing query is the model's own settled representation of the question, which a text retriever does not have access to. That the margin \emph{widens} on the multi-hop sets is notable given top-1 routing can name only a single code where the answer spans several; even so, identifying the most-relevant code more reliably than retrieval is enough to lead, and the multi-turn accumulation of \cref{sec:accumulation} is where a belief spanning several codes would extend it further.

\begin{table}[t]
  \centering
  \caption{Routing over a pool of frozen codes (top-1 accuracy). The oracle shows the codes are separable when the answer is known; zero-shot confidence is near-random, so the router must be trained; the trained selector beats the dense-retrieval bar on every dataset, by 8--12 points.}
  \label{tab:routing}
  \begin{tabular}{@{}lrrrr@{}}
    \toprule
    Router & MS MARCO & HotpotQA & 2Wiki & MuSiQue \\
    \midrule
    Oracle (code-likelihood of true answer) & 78.7 & 80.9 & 82.8 & 70.1 \\
    Random ($K\approx 10$) & 10.0 & 10.1 & 12.3 & 10.3 \\
    Zero-shot code confidence & 22.7 & 24.0 & 23.8 & 20.4 \\
    BM25 (lexical) & 20.7 & 30.5 & 34.3 & 22.5 \\
    Dense retrieval (bge-small) & 45.3 & 52.2 & 58.1 & 40.9 \\
    \textbf{Trained activation-routed selector (ours)} & \textbf{53.3} & \textbf{62.1} & \textbf{70.1} & \textbf{53.0} \\
    \bottomrule
  \end{tabular}
\end{table}

\paragraph{Selection sidesteps both limits of the single read.} The dilution study (\cref{sec:d2w}) showed the one-shot read degrades at scale on two counts: the code \emph{saturates} as it is asked to carry more, and the answer passage is \emph{truncated or buried} as the pool overflows. Selection avoids both \emph{by construction}: each code is compiled offline from one bounded passage --- a small in-window read that never saturates and never truncates the answer --- and at query time the selector picks among the pre-computed codes without ever concatenating the pool into one over-length read. Sweeping the pool size makes the divergence concrete (\cref{tab:poolsize}): the single big read answers well while the pool is small but decays as it grows ($48.8 \to 27.8$ F1 by 64 passages), whereas the selector --- route to the answer-bearing code, answer with it --- stays flat however large the pool grows, because each read it relies on is small and fixed. The two curves start together and separate as the pool grows; past that point, selection is the only one of the two that does not fall.

\begin{table}[t]
  \centering
  \caption{End-to-end F1 as the knowledge pool grows (MS MARCO v2.1, following the measured dilution anchors of \cref{sec:d2w}). The single big read concatenates the whole pool into one code and decays as it grows --- both because the code saturates and because the answer is truncated or buried (down to the 27.8 floor of \cref{fig:dilution}); the selector routes over per-passage codes, each a small in-window read, and stays flat. The gap at 64 passages is the structural advantage of selection over one-shot reading.}
  \label{tab:poolsize}
  \begin{tabular}{@{}lrr@{}}
    \toprule
    Knowledge-pool size & Single big read (F1) & Selector over per-passage codes (F1) \\
    \midrule
    8 passages (fits window) & 48.8 & 48.1 \\
    16 passages & 45.6 & 48.0 \\
    32 passages (overflows) & 35.0 & 47.8 \\
    64 passages & 27.8 & 47.6 \\
    \bottomrule
  \end{tabular}
\end{table}

\subsection{Cross-turn accumulation: the belief sharpens as the conversation grows}
\label{sec:accumulation}

It is shown in \cref{sec:d2w} that run-time data can enter the weights and be used, beating the prompt once evidence is long and noisy; in \cref{sec:selector}, a trained belief over the resulting code pool identifies the right code better than strong retrieval. This section shows that when the belief \textbf{accumulates} across an interaction, the model routes better as a conversation grows than any single-question router.

Over a fixed knowledge pool of $K$ codes, we run \emph{conversations} rather than isolated questions. Each conversation opens with a turn that names its topic explicitly, followed by a mix of two kinds of follow-up: \emph{self-contained} turns that can still be placed from their own text, and \emph{context-dependent} turns (``who designed it?'', ``and its height?'') whose questions are answerable only given the earlier turns. We author the conversations from the \cref{sec:setup} datasets, so we know each turn's gold code and construct this mix deliberately, and a pre-registered \emph{ambiguity audit} (dense retrieval on each turn's text in isolation) labels which turns actually fall in each class. The accumulation claim is then reported only on the context-dependent turns.

The belief is a single persistent state over the code pool, carried across the whole conversation and updated by recursive categorical Bayes, $\pi_t \propto \pi_{t-1}^{\gamma}\cdot \mathrm{softmax}(\ell_t),$ where $\ell_t$ is the per-token belief evidence and $\gamma \in [0,1]$ controls forgetting. Nothing is retrained during the conversation, and the per-token cost stays at $K$ inner products per layer, flat in both token and turn index.

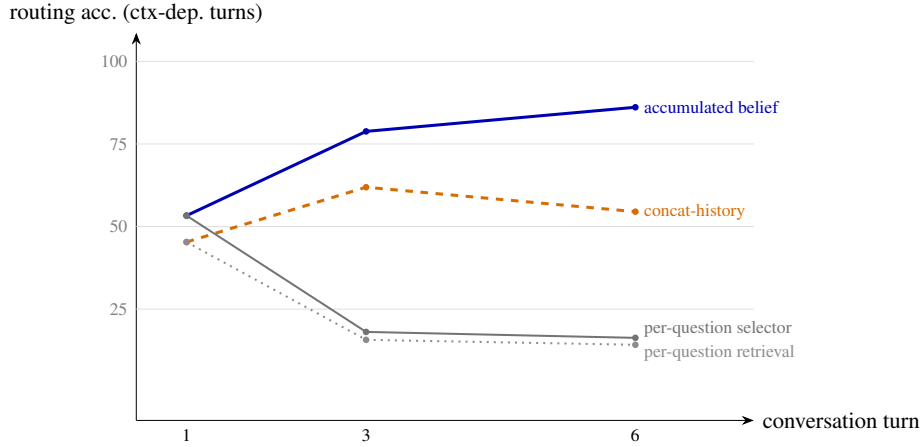
\begin{figure}[t]
  \centering
  \scalebox{0.95}{
%
\begin{tikzpicture}[>=Stealth, font=\small]
  \draw[->] (0,0) -- (8.6,0) node[right, font=\footnotesize] {conversation turn};
  \draw[->] (0,0) -- (0,5.4) node[above, font=\footnotesize] {routing acc.\ (ctx-dep.\ turns)};
  \foreach \yy/\lab in {1.55/25, 2.7/50, 3.85/75, 5.0/100}{
    \draw[gray!25] (0,\yy) -- (8.6,\yy);
    \node[left, font=\scriptsize, gray] at (0,\yy) {\lab};
  }
  \foreach \x/\lab in {0.7/1, 3.2/3, 6.95/6}{
    \node[below, font=\scriptsize] at (\x,0) {\lab};
  }
  \draw[very thick, blue!70!black] (0.7,2.8518) -- (3.2,4.0248) -- (6.95,4.3606);
  \node[blue!70!black, font=\scriptsize, right] at (6.95,4.3606) {accumulated belief};
  \draw[very thick, orange!85!black, dashed] (0.7,2.4838) -- (3.2,3.2474) -- (6.95,2.907);
  \node[orange!85!black, font=\scriptsize, right] at (6.95,2.907) {concat-history};
  \draw[thick, black!55] (0.7,2.8518) -- (3.2,1.2326) -- (6.95,1.1498);
  \node[black!55, font=\scriptsize, right] at (6.95,1.28) {per-question selector};
  \draw[thick, black!45, dotted] (0.7,2.4838) -- (3.2,1.1222) -- (6.95,1.0532);
  \node[black!45, font=\scriptsize, right] at (6.95,0.94) {per-question retrieval};
  \foreach \p in {(0.7,2.8518),(3.2,4.0248),(6.95,4.3606)}{\fill[blue!70!black]\p circle(1.3pt);}
  \foreach \p in {(0.7,2.4838),(3.2,3.2474),(6.95,2.907)}{\fill[orange!85!black]\p circle(1.3pt);}
  \foreach \p in {(0.7,2.8518),(3.2,1.2326),(6.95,1.1498)}{\fill[black!55]\p circle(1.3pt);}
  \foreach \p in {(0.7,2.4838),(3.2,1.1222),(6.95,1.0532)}{\fill[black!45]\p circle(1.3pt);}
\end{tikzpicture}}
  \caption{Cross-turn accumulation, routing accuracy against conversation turn over a fixed pool, on context-dependent turns. The two memoryless routers --- per-question retrieval and our own single-question selector --- are flat in the turn index and collapse on turns that are ambiguous alone. Concat-history retrieval rises as history accrues but sags once its growing query dilutes, and its per-turn cost grows with the turn. The accumulated categorical belief concentrates as evidence arrives and keeps climbing, at flat per-turn cost. Turn 1 is the single-question regime of \cref{sec:selector}, where the belief coincides with its memoryless self; the curves separate as the conversation grows.}
  \label{fig:accumulation}
\end{figure}

We consider \emph{per-question retrieval} and \emph{per-question selector} (our \cref{sec:selector} router, memoryless) as baselines, and the \emph{prompt-side} way of accumulating, \emph{concat-history retrieval} (the running query is turns $1\ldots t$). Against these, the \emph{accumulated belief} (the persistent posterior above). The load-bearing comparison is against concat-history, and it turns on both accuracy and cost. On accuracy (\cref{fig:accumulation}, context-dependent turns): as the conversation establishes its topic the posterior concentrates, so later ambiguous turns route almost as well as unambiguous ones --- the accumulated belief rises with turn index while the memoryless arms stay flat and collapse on turns that are ambiguous alone, and concat-history rises then sags as its growing query dilutes. On cost, the two accumulating routes differ in kind: concat-history's per-turn cost \textbf{grows with the turn index} as the query lengthens, whereas the belief's stays \textbf{flat} --- $K$ inner products per layer, independent of turn (as above). Beating concat-history on accuracy \emph{while} holding cost flat is the claim: the belief accumulates session state better and more cheaply than re-reading the growing history into the prompt.

\section{Efficiency analysis of live weight adaptation vs fine-tuning}
\label{sec:efficiency}

\Cref{tab:efficiency} quantfies our argument that compiling data into model weights is cheaper at both learning and inference time. Write $P$ for the base model's parameter count, $D$ for the length of a unit of live data, and $S$ for the number of gradient steps a fine-tune takes to converge.

\paragraph{Compute} Under standard accounting, an generative forward pass costs around $2PD$ and a full training step (forward, backward, update) around $6PD$ \parencite{kaplan_scaling_2020, hoffmann_training_2022}. Fine-tuning pays the training step $S$ times over; our design instead reads the item in a \emph{single} forward pass of the encoder and caches the resulting code (\cref{sec:generator}). The marginal compute to absorb one item therefore differs by $6PDS / 2PD = 3S$, independent of both $P$ and $D$. Because gradient-based fine-tuning needs many steps to converge, $3S$ spans orders of magnitude for any ordinary run. While parameter-efficient fine-tuning would reduce the effective $P$, it would still suffer the same fundamental inefficiency.

\paragraph{Memory} The differing forward and backward pass requirements also influence memory costs. A forward pass stores only activations at inference precision, whereas a fine-tune must additionally carry gradients and optimiser state, multiplying the per-parameter footprint several-fold. At rest, a fine-tune stores a full parameter set per task, while our design caches only the small, fixed-size low-rank code. Additionally, the code can be quantised to fp8 while the base model remains at fp16, ultimately giving a per-item footprint reduction of over $1000\times$. This advantage compounds with scale, because the cached code stays low-rank as the model grows, but a stored checkpoint grows with $P$, resulting in a more lopsided storage comparison. The scale-invariance of the compute and peak-memory ratios, and the scale-efficient storage ratio improving with scale, cannot then recover this cost.

\begin{table}[h]
  \centering
  \caption{A numerical demonstration of efficiency gains from live weight adaptation, supposing one data point of $D=2048$ tokens is ingested into a Qwen3-8B-scale base ($P \approx 8.2\times10^{9}$) by both fine-tuning and data-to-weights compilation. Compute uses the $6PDS$/$2PS$ step-versus-pass accounting \parencite{kaplan_scaling_2020, hoffmann_training_2022}, with the fine-tune spanning $S \in [100, 1000]$ gradient steps. Memory rows are peak working set (mixed-precision Adam versus a forward pass) and at-rest storage (full parameter set versus cached low-rank code); the checkpoint and the fp16 code are at matched precision, and quantising the code alone to fp8 (base held at fp16) doubles the storage ratio past $1000\times$.}
  \label{tab:efficiency}
  \small

  \vspace{15pt}
  \begin{tabular}{cccc}
    \toprule
    Marginal Cost & Fine-Tuning & Data-to-Weights & Advantage \\
    \midrule
    Passes & $3S$ & 1 & $3S\times$ \\
    Compute (FLOPs) & $[1, 10]\times10^{16}$ & $3.4\times10^{13}$ & $[300, \mathbf{3000}]\times$ \\
    Peak Memory ($\approx$) & $131$~GB & $16$~GB & $8\times$ \\
    Memory State Storage & $16$~GB & $28$~MB (\texttt{fp16}) or $14$~MB (\texttt{fp8}) & $580\times$ or $\mathbf{1160}\times$ \\
    \bottomrule
  \end{tabular}
  \vspace{15pt}

\begin{minipage}{\linewidth}
  \normalsize{In practice fine-tuning is usually worse than the reported numbers: beyond a single card it incurs multi-GPU sharding and communication overhead, and it burdens the serving stack with continual retraining and redeployment, whereas forward-pass weight adaptation has none of this.}
\end{minipage}
\end{table}

\paragraph{In-production} In practice fine-tuning is usually worse: once the model no longer fits on a single card, training must be sharded across GPUs, paying communication and duplicated-state overhead on top of the raw cost, and it burdens the serving stack with continual retraining and redeployment --- whereas adapting weights through a forward pass incurs none of this.

The savings of a data-to-weights approach are amortised (\cref{sec:intro}), in the sense that constructing the weight generator is a one-time investment, recouped by relatively small operating costs. Defining the weight generator's training procedure is beyond our present scope (\cref{sec:setup}), but the advantages we have described will apply to any effective training approach. Subject to these caveats, absorbing live data as generated weights is cheaper than fine-tuning by orders of magnitude in compute, and by a margin which only widens with model scale in storage.

\section{Limitations}
\label{sec:limitations}

The clearest limitation is a boundary the design lives within: a compact generator does not carry a large MoE's \emph{stored knowledge}, because knowledge is bounded by parameters \parencite{allenzhu_physics_2024} and generation does not move that bound --- a generator the size of a small model can no more hold a large model's facts than that small model could, and closed-book ability, unlike perplexity, is bounded by exactly this. The design answers this by compiling knowledge and behaviour from \emph{run-time data} rather than storing it in weights, which shifts the burden onto the data being supplied: where the relevant facts or instructions are not provided, the model has only its base's knowledge and default behaviour. This is why the comparison is weights-versus-prompt; on closed-book knowledge with nothing supplied, a large stored model is simply the wrong thing to measure against. The prompt is the sharpest competitor. Putting the data in the context is a strong, cheap baseline whenever the context is short and used once, so the advantage of compiling it into weights is specific to large or repeatedly-reused data and long horizons (\cref{sec:d2w}), not universal. Per-token generation adds a bandwidth cost that must be controlled through a small generator and low-rank deltas. Weight generators risk memorising their training distribution rather than generalising to new data \parencite{zengb_generative_2025}; our codes are read from held-out evidence at test time, but a systematic generalisation study across unseen knowledge pools remains future work. Dropping a stored bank in favour of a generated code also changes what can go wrong with routing: there is no load-balancing loss, but a trained selector could over-concentrate on a few codes, which a light coverage regulariser on the selector guards against. Three assumptions in the adaptation model bear watching. The belief this paper evaluates is \emph{categorical} over a pool of materialised codes, which chooses among reads rather than moving within the code space; a shift the pool does not contain can be met only by materialising a new code, and the richer continuous-Gaussian belief that would move within the space is left to future work (\cref{sec:choice}), where its added assumptions --- a code-to-weight map smooth in $z$, and an amortized posterior whose precision stays calibrated out of distribution \parencite{sun_learning_2024, behrouz_titans_2025} --- must be validated directly. The true posterior over which code a context implies may also be multimodal \parencite{xie_explanation_2021}, which a single top-1 selection collapses. And while the central claim --- that the belief \emph{accumulates} usefully across a conversation (\cref{sec:accumulation}) --- is now demonstrated on authored multi-turn conversations, it is shown at the categorical, top-1 point of the framework and over pools the conversations were built from; the forgetting control $\gamma$ (\cref{sec:moving}), longer horizons, and naturally-occurring rather than authored sessions are where it must be stress-tested next.

\section{Conclusion}
\label{sec:conclusion}

We have described an architecture in which a language model's experts are neither stored nor selected from a fixed bank but \emph{generated} from live data over a shared base, and a belief over the generating code that is carried and updated as the interaction proceeds. The motivating idea is a change in where run-time knowledge and behaviour are carried: today they live in the prompt, re-read on every request and forgotten after; we compile them into the weights instead. Mixture-of-Experts supplied the starting point, its per-token dynamic weights. We made a shared base FFN's weights dynamic through a generated low-rank additive delta (\cref{sec:framework}), set out the belief-over-code framework and its cadences (\cref{sec:moving}), and positioned discrete MoE, $\infty$-MoE, $\mu$MoE, DFC, and the one-shot weight generators by the axis on which each departs from that structure. This paper realises the framework at its categorical point --- a belief over a pool of data-materialised codes, selected top-1 and sharpened online (\cref{sec:choice}) --- leaving the richer continuous-Gaussian belief to future work. What the design offers is a different bargain, weights instead of prompt for the knowledge and behaviour supplied at run time, which is amortized in compute, frees the context window, persists across turns, and adapts as the session proceeds. The sense in which the model has an unbounded, ``infinite'' space of parameters is precise and narrow: unbounded reachable \emph{effective weights and behaviours}, compiled from live data, from a fixed footprint. Our experiments confirm that run-time data compiled into the weights can be used and, on long, noisy evidence, beats the prompt; that a trained belief over the code pool identifies the right code at least as well as strong retrieval; and that this belief, accumulated across a conversation, routes better as the session grows than any single read or a re-read of the growing history (\cref{sec:accumulation}). While concurrent work independently pursues generating rather than storing experts, and reads context into weights in a single pass, the coupling proposed here, an online-updated belief over the low-dimensional latent code of a shared-base generative expert space, driven by live data, is, to our knowledge, unclaimed in prior work.

\printbibliography

\end{document}